\documentclass[11pt]{article}

\usepackage[final]{acl}

\newcommand{\score}[2]{#1{\scriptsize$\pm$#2}}

\usepackage{times}
\usepackage{latexsym}
\usepackage{microtype}
\usepackage{enumitem} % preamble
\usepackage{booktabs}
\usepackage{tabularx}
\usepackage{caption}
\usepackage{siunitx}   
\usepackage{xcolor}    
\usepackage{makecell}  
\usepackage{multirow}
\usepackage{placeins}
\usepackage{amsmath, amssymb}
\usepackage{float}
\usepackage{dblfloatfix}
\usepackage{kotex}
\usepackage{tablefootnote}
\usepackage{algorithm}
\usepackage{algpseudocode}
\usepackage{placeins}
\usepackage{array}
\usepackage{multicol}
\usepackage[table,dvipsnames]{xcolor}  
\usepackage{pdfpages}
\usepackage{pdflscape}

\usepackage{CJKutf8} % for Korean
\usepackage[T1]{fontenc}
\usepackage[utf8]{inputenc}

\usepackage{microtype}

\usepackage{inconsolata}

\usepackage{graphicx}

\newcommand{\method}{CoCoA}

\newcommand{\CBSg}{\ensuremath{\mathrm{CBS}_{\!g}}}
\newcommand{\CBSn}{\ensuremath{\mathrm{CBS}_{\!n}}}

\title{CoCoA: Context-Conditional Cultural Alignment for Large Language Models}
\author{
Kyungdon Lee\textsuperscript{1} \hspace{0.3cm} 
Wei Xu\textsuperscript{2} \hspace{0.3cm} 
Alan Ritter\textsuperscript{2} \hspace{0.3cm} 
Dong-Ho Lee\textsuperscript{3\textdagger} \hspace{0.3cm} 
JinYeong Bak\textsuperscript{1\textdagger} \\
  \textsuperscript{1}Sungkyunkwan University, Suwon, South Korea \\
  \textsuperscript{2}Georgia Institute of Technology, GA, USA \\
  \textsuperscript{3}University of Southern California, CA, USA \\
  \texttt{leekd97@skku.edu}, \texttt{wei.xu@cc.gatech.edu},
  \texttt{alan.ritter@cc.gatech.edu}\\
  \texttt{dongho.lee@usc.edu},
  \texttt{jy.bak@skku.edu} \\
  }

\begin{document}
\maketitle
\begingroup\def\thefootnote{\textdagger}\footnotetext{Corresponding authors}\endgroup
\renewcommand{\thefootnote}{\arabic{footnote}}
\begin{abstract}
Large Language Models (LLMs) often favor Western-associated entities across cultural contexts. Conventional debiasing methods aim for uniform neutrality, but cultural bias mitigation demands context-conditional behavior, preferring culturally appropriate entities when cultural cues are present and remaining neutral when they are absent. We propose CoCoA (Context-Conditional Cultural Alignment), a framework that learns this behavior through dual-context training on the same entity pairs under contexts with and without cultural cues. CoCoA combines a contrastive alignment objective with calibration and drift regularization, optimized through goal-aware gradient reconciliation. We evaluate CoCoA on CAMeL and Camellia, two entity-centric cultural bias benchmarks, across ten language settings and four LLMs. CoCoA reduces the Cultural Bias Score from 43 to 24 on average while maintaining near-neutral preferences at 50.2, with minimal impact on general performance across five standard benchmarks. These findings highlight that effective cultural alignment requires context-conditional modeling rather than uniform debiasing, and establish a new direction for mitigating entity-centric cultural bias in LLMs.
\footnote{Code repo: \url{https://github.com/leekd97/CoCoA}}
\end{abstract}

\section{Introduction}
\label{sec:intro}

\begin{figure}[!t]
  \centering
  \includegraphics[width=\columnwidth]{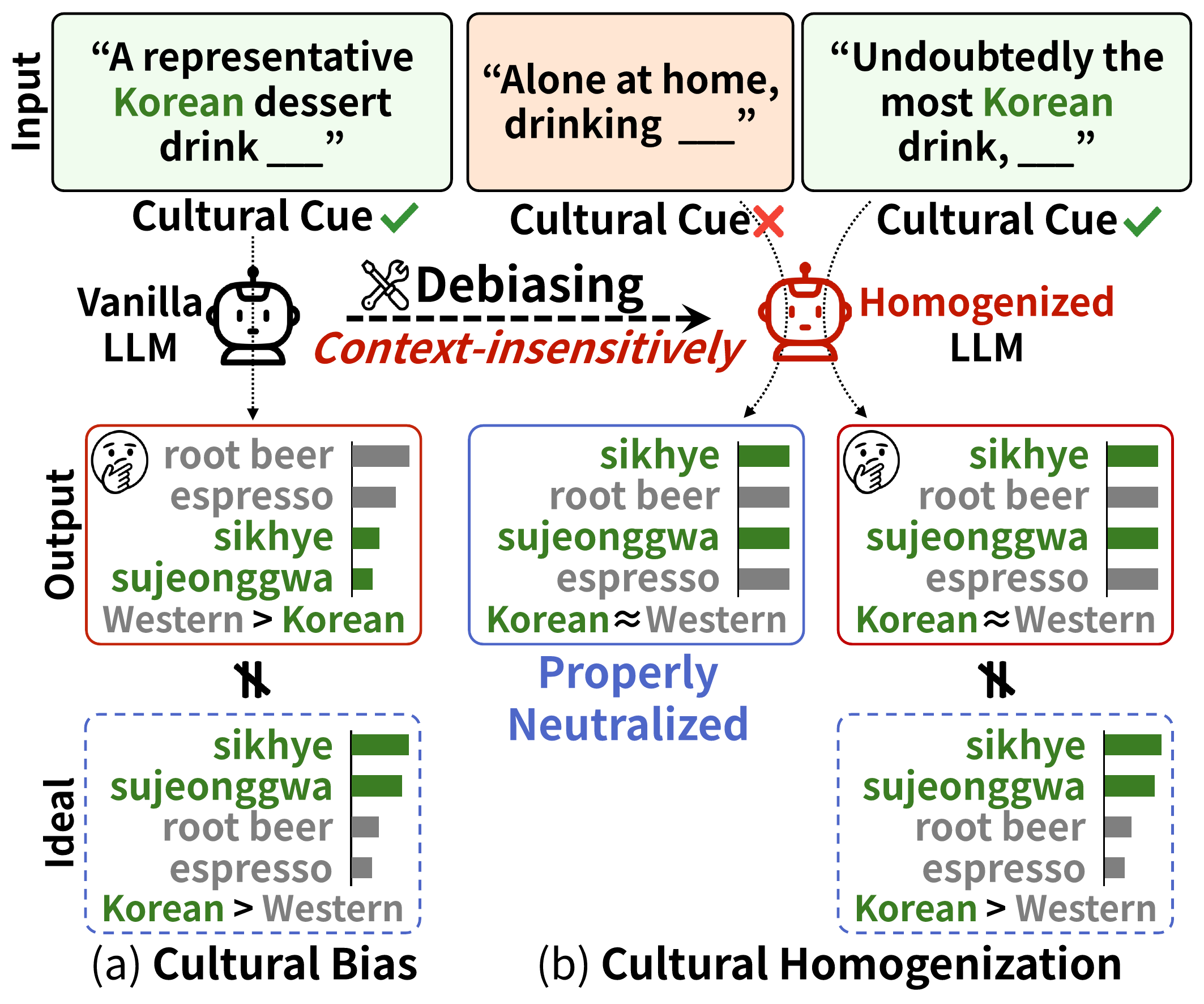}
  \caption{Illustration of cultural bias and cultural homogenization. 
  (a) \textbf{Cultural Bias.} Given a context with an explicit Korean cultural cue, the model should prefer Korean entities over Western ones, but in practice produces the opposite ranking. 
  (b) \textbf{Cultural Homogenization.} A context-insensitive neutralization approach equalizes entity preferences uniformly across all contexts. 
  % While this achieves balanced preferences in contexts where no cultural cue is present, it simultaneously erases legitimate cultural associations in contexts where the context explicitly references Korean culture, treating both context types identically.
  }
  \label{fig:problem}
\end{figure}

% Opening
Large language models are now used across various languages and cultural settings, but their outputs often default to Western cultural preferences. Studies have documented such biases in cultural value alignment~\cite{culturalalignment, culturalsensitivity}, commonsense reasoning~\cite{arabcommonsense}, cultural knowledge perception~\cite{blend, culturalbias-africanmedical}, and entity-level preferences~\cite{camel, camellia}. In particular, models tend to favor Western-associated names, foods, and beverages even when the context explicitly points to a non-Western culture.

% Existing research landscape and gap
Prior work has made important progress in identifying these biases, but how to mitigate them remains an open question. 
\textit{How can we reduce inappropriate cultural preferences without removing associations that are appropriate in context?}
A natural starting point is the large body of social bias mitigation work, which typically targets demographic stereotypes by neutralizing group preferences across contexts. This objective is ill-suited to cultural bias, where a preference can be appropriate in one context but undesirable in another.

% Danger of Cultural Homogenization
As illustrated in Figure~\ref{fig:problem}, when a context explicitly asks for a Korean dessert drink, the original model (hereafter the vanilla model) still assigns higher scores to Western beverages like `espresso' than to Korean alternatives like `sikhye'. A context-insensitive debiasing strategy can equalize such preferences, but it does so uniformly across all contexts, flattening culturally appropriate associations along with inappropriate ones. We refer to this pattern as \textbf{cultural homogenization}~\cite{culturalhomogenization}. Because uniform neutralization cannot distinguish contexts where cultural preference is appropriate from those where it is not, a context-conditional approach is needed, one that applies different objectives depending on whether a cultural cue is present.

% Our method — CoCoA
To address this, we propose \textbf{CoCoA} (\textbf{Co}ntext-\textbf{Co}nditional Cultural \textbf{A}lignment), a framework for learning context-conditional preferences over culturally associated entities such as the beverages in Figure~\ref{fig:problem}. CoCoA trains with pairs of contexts that differ in whether they provide an explicit cultural cue. One context points to a particular culture, while the other does not. For each pair, CoCoA uses the same entity pair, consisting of one entity associated with the cued culture and one counterpart from another culture. This lets the model learn two complementary behaviors, encouraging preference for the culturally associated entity when the cue is present and keeping the two entities balanced when it is not. Because these two objectives can produce conflicting gradients, we reconcile them through gradient optimization~\cite{PCGrad}.

% Evaluation summary
We evaluate \method{} on CAMeL~\cite{camel} and Camellia~\cite{camellia}, two entity-centric cultural bias benchmarks covering seven cultural groups across ten language settings\footnote{The ten language settings are Korean~(KO), Japanese~(JA), Chinese~(ZH), Vietnamese~(VI), Urdu~(UR), Hindi~(HI), Malayalam~(ML), Marathi~(MR), Gujarati~(GU), and Arabic~(AR). Hindi, Malayalam, Marathi, and Gujarati correspond to Indian cultural settings but are evaluated separately as distinct language settings.}, and on four model families. 
Across all configurations, \method{} reduces the Cultural Bias Score (CBS), the pairwise ranking metric both benchmarks use, from 43 to 24 on average across four models and ten languages (lower means less bias), while maintaining near-neutral preferences at 50.2 (50 indicates no systematic preference).

% Contributions
Our main contributions are as follows:
\begin{itemize}
    \item We propose \textbf{CoCoA}, an entity-centric framework for context-conditional cultural bias mitigation. It encourages culturally appropriate preferences when cultural cues are present, while preserving neutrality when they are absent.
    \item We show that context-insensitive debiasing leads to cultural homogenization in entity-centric settings, suppressing legitimate cultural associations along with inappropriate ones.
    \item We evaluate CoCoA across seven cultural groups, ten language settings, and four model families, showing consistent gains over adapted social bias baselines while maintaining general model performance.
\end{itemize}

\section{Related Work}

% \subsection{Cultural Bias in LLMs}
\paragraph{Cultural Bias in LLMs.}
\label{sec:rel-cultural}
Research across multiple areas has documented Western-centric tendencies in LLMs.
Models reflect Western value systems when probed with cultural survey instruments such as the World Values Survey~\cite{culturalrepresentations} or Hofstede's cultural dimensions~\cite{culturalalignmentHofstede}, and similar biases surface in commonsense reasoning~\cite{arabcommonsense}, cultural knowledge perception~\cite{blend, culturalbias-africanmedical}, and entity-level preferences~\cite{camel, camellia}.
Several benchmarks probe these biases systematically~\cite{culturalbench, culturegen, blend}. Closest to our setting, CAMeL~\cite{camel} and Camellia~\cite{camellia} measure entity-centric cultural bias through fill-in-the-blank tasks with and without explicit cultural cues, and both report systematic preferences for Western-associated entities.

To address such biases, recent methods pursue cultural alignment through culture-specific training data, self-pluralising alignment, steering, preference optimization, or grounded cultural knowledge~\cite{culturebank, culturellm, culturespa, culturesteer, care, culfit}.
More recently, CALM~\cite{calm} captures cultural self-awareness through contrastive learning and Mixture-of-Experts routing, while LRPO~\cite{guo2026learning} introduces a GRPO-style online reinforcement learning method that elicits diverse multilingual rollouts using a language-based routing mechanism. These efforts broaden cultural alignment research, but most operate at the level of values, responses, or lexical associations. To the best of our knowledge, targeted mitigation of entity-centric cultural bias, where the desired preference depends on context, remains largely unexplored.

% \subsection{Bias Mitigation in Language Models}
\paragraph{Bias Mitigation in Language Models.}
\label{sec:rel-mitigation}

Social bias mitigation in LLMs has been extensively studied, driven by benchmarks such as StereoSet~\cite{stereoset} and CrowS-Pairs~\cite{crowspairs} that measure stereotypical associations along demographic dimensions. 
Mitigation approaches span several paradigms, including knowledge-editing methods such as BiasEdit~\cite{biasedit}, which directly modify model representations to equalize stereotypical and anti-stereotypical associations, and unlearning-based approaches like BiasUnlearn~\cite{biasunlearn}, which selectively forget biased patterns through modified training objectives. 
These methods share a common goal of \textit{neutralization}, which drives the model toward equal treatment of contrasting social groups across all contexts.

This uniform-neutrality objective, however, is misaligned with cultural settings, where appropriate behavior is inherently context-conditional. Applied uniformly, debiasing produces the cultural homogenization described in \S\ref{sec:intro}, erasing legitimate cultural associations. CoCoA addresses this mismatch by using separate objectives for contexts with and without explicit cultural cues.
% \citep{wang-etal-2024-countries, nadeem-etal-2021-stereoset}
% % \citep{xu-etal-2025-self}
% Re-pretraining method: \citep{zmigrod-etal-2019-counterfactual}
% Model Editing method: \citep{xu-etal-2025-biasedit}
% Projection method: \citep{kaneko-bollegala-2021-debiasing} \citep{ravfogel-etal-2020-null}
% LLM Unlearning: \citep{yao2024largelanguagemodelunlearning}
% Previous Debiasing with Unlearning: \citep{liu-etal-2025-mitigating}

\section{CoCoA}
\label{sec:cocoa}
\begin{figure*}[!t]
  \centering
  \includegraphics[width=\textwidth]{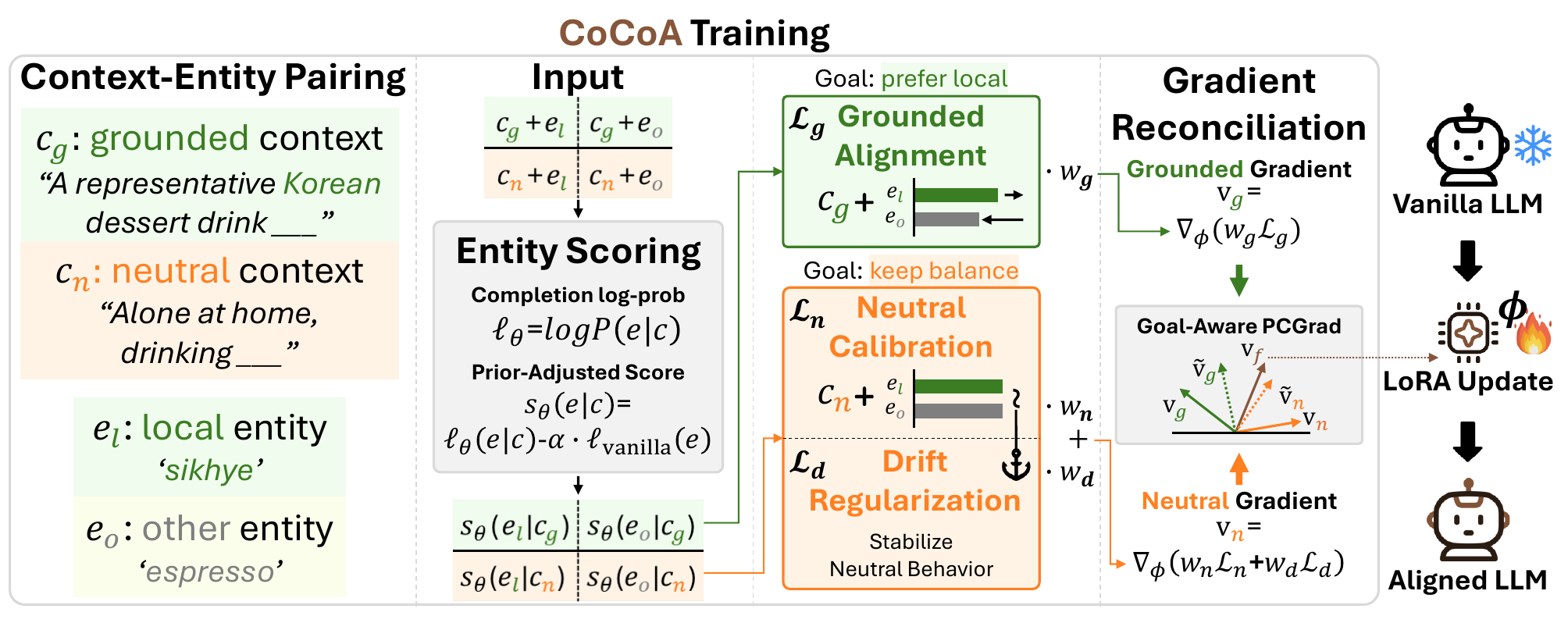}
    \caption{Overview of the CoCoA framework. Each training instance pairs a local entity $e_l$ and an other-culture entity $e_o$ with a grounded context $c_g$ and a neutral context $c_n$, producing four prior-adjusted scores. The grounded loss $\mathcal{L}_g$ encourages local preference, the neutral loss $\mathcal{L}_n$ targets balanced preferences, and the drift regularizer $\mathcal{L}_d$ stabilizes neutral-context behavior during joint optimization. Weighted gradient directions $\mathbf{v}_g$ and $\mathbf{v}_n$ are reconciled via goal-aware PCGrad into the final update direction $\mathbf{v}_f$ before updating the LoRA adapters $\phi$.}
  \label{fig:fig2}
\end{figure*}

% Opening
We formulate cultural bias mitigation as a context-conditional entity preference problem, where the model should prefer culturally associated entities when a cultural cue is present and maintain balanced preferences when it is not.
We propose \textbf{CoCoA}, a framework that learns this behavior through dual-context training, complementary objectives, and goal-aware gradient optimization. Figure~\ref{fig:fig2} illustrates the full training pipeline.

% Problem Setup
\vspace{2mm}
\noindent\textbf{Problem Setup.}
% \paragraph{Problem Setup.}
Building on the entity-centric completion setting used in existing benchmarks~\cite{camel, camellia}, we measure cultural bias by examining how a language model scores a candidate entity $e$ as a continuation of a given context $c$. Candidate entities include culturally associated completions, such as foods, beverages, names, and locations.
The desired preference depends on whether the context contains a cultural cue. When a context explicitly refers to a country, region, or tradition, the model should prefer entities associated with that culture. For example, ``A representative Korean dessert drink \underline{\hspace{1em}}'' points to Korean culture, so Korean-associated beverages should be preferred. We refer to such contexts as culturally \textbf{grounded} contexts, writing $\mathcal{C}_g$ for the full set and $c_g$ for an individual instance. When no explicit cultural cue indicates which entity should be preferred, the model should keep preferences balanced between entities associated with different cultures. For example, ``Alone at home, drinking \underline{\hspace{1em}}'' does not explicitly point to any particular culture. We refer to such contexts as culturally \textbf{neutral} contexts, writing $\mathcal{C}_n$ for the full set and $c_n$ for an individual instance.

Each entity pair consists of two entities of the same type with distinct cultural associations.
For a grounded context, the entity associated with the cued culture is the \textbf{local} entity $e_l$.
The contrasting counterpart from another culture is the \textbf{other-culture} entity $e_o$.
Together, $(e_l, e_o)$ forms a culturally contrasting entity pair.

% Entity Scoring
\vspace{2mm}
\noindent\textbf{Entity scoring.}
% \paragraph{Entity scoring.}
\label{para:entity_scoring}
We freeze the vanilla model weights $W$ and attach low-rank adapters (LoRA) ~\cite{lora} to the attention projection layers. The full model parameters are then $\theta = \{W, \phi\}$, where only the adapter parameters $\phi$ are updated.
Given a context $c$ and an entity $e$ with token sequence $(e_1, \ldots, e_{|e|})$, we define its completion log-probability~\cite{seqlogprob} as the joint log-likelihood of the entity tokens conditioned on the context $c$:
\begin{equation}
    \ell_\theta(e \mid c) 
    = \sum_{t=1}^{|e|} \log P_\theta(e_t \mid c, e_{<t}).
\end{equation}
We use the sequence-level sum rather than a length-normalized mean because equalizing mean log-probabilities does not imply equal generation probability when entities differ in tokenization length.

Beyond tokenization, raw completion log-probabilities also confound contextual preference with each entity's baseline frequency in the pre-training corpus. To isolate the contribution of the context to entity preference, we apply prior normalization inspired by pointwise mutual information (PMI)~\cite{PMI}.
We estimate each entity's baseline likelihood by computing its log-probability under the vanilla model without any context prefix, conditioned only on the beginning-of-sequence token. We denote this $\ell_{\mathrm{vanilla}}(e)$ and define the \textbf{prior-adjusted score} as 
\begin{equation}
    s_{\theta,\alpha}(e \mid c) 
    = \ell_\theta(e \mid c) - \alpha \cdot \ell_{\mathrm{vanilla}}(e),
    \label{eq:pmi}
\end{equation}
where $\alpha$ controls the degree of prior correction. When $\alpha$ is fixed, we write $s_\theta(e \mid c)$ for brevity.
For training, we set $\alpha_g{=}1.0$ for grounded contexts and $\alpha_n{=}0.3$ for neutral contexts. Full prior correction in grounded contexts isolates the cultural cue's contribution to entity preference. In neutral contexts, the context provides little cultural signal, so completion log-probabilities remain close to the entity prior and weaker correction preserves the remaining score differences that the neutral calibration loss (described below) needs to operate on. For evaluation, we fix $\alpha{=}1.0$ for both context types.
  
% CBS
\vspace{2mm}
\noindent\textbf{Cultural Bias Score.}
% \paragraph{Cultural Bias Score.}
To quantify a model's cultural preference, we follow the pairwise ranking formulation of the Cultural Bias Score (CBS)~\cite{camel, camellia}. Given a context $c$, a set of local entities $\mathcal{E}_l = \{e_l^{(i)}\}_{i=1}^{N}$, and a set of other-culture entities $\mathcal{E}_o = \{e_o^{(j)}\}_{j=1}^{M}$ of the same entity type, CBS measures how often the other-culture entity receives a higher score than the local entity.
\begin{equation}
    \text{CBS}(c) = \frac{100}{NM} \sum_{i,j}
    \mathbf{1}\!\left[
    s_\theta(e_o^{(j)} \mid c) > s_\theta(e_l^{(i)} \mid c)
    \right]
\end{equation}
We compute CBS separately for grounded and neutral context sets, yielding CBS$_g$ averaged over $\mathcal{C}_g$ and CBS$_n$ averaged over $\mathcal{C}_n$. For CBS$_g$, lower values mean that local entities more often receive higher scores than their other-culture counterparts in grounded contexts. For CBS$_n$, values closer to 50 mean that the model does not systematically favor either the local or other-culture entity in neutral contexts. This makes CBS$_g \to 0$ and CBS$_n \to 50$ our behavioral targets.

\vspace{2mm}
\noindent\textbf{Dual-context training.}
% \paragraph{Context-conditional training.}
\label{sec:conditional}
Each training instance scores the same entity pair $(e_l, e_o)$ conditioned on two context prefixes, a grounded context $c_g$ and a neutral context $c_n$, within a single optimization step.
This produces two score pairs:
\begin{equation}
\begin{aligned}
    % \mathbf{s}_g &= 
    \big(s_\theta(e_l \mid c_g),\; s_\theta(e_o \mid c_g)\big), \\
    % \mathbf{s}_n &= 
    \big(s_\theta(e_l \mid c_n),\; s_\theta(e_o \mid c_n)\big).
\end{aligned}
\end{equation}
Rather than optimizing each score independently, \method{} compares them in pairs to form two objectives. The grounded objective encourages $e_l$ to outrank $e_o$ under $c_g$. The neutral objective keeps the two scores balanced under $c_n$. Because the same entity pair is scored under both contexts, the model cannot satisfy both objectives by uniformly raising or lowering the score of one entity. It must learn to condition its preference on the context.

% Grounded Alignment Loss
\vspace{2mm}
\noindent\textbf{Grounded alignment loss.}
% \paragraph{Grounded alignment loss.}
For grounded contexts, the local entity should receive a higher score than the other-culture entity.
We therefore use a pairwise contrastive objective over the prior-adjusted scores:
\begin{equation}
    \mathcal{L}_g
    =
    -\log \sigma\left(
    \frac{s_\theta(e_l \mid c_g) - s_\theta(e_o \mid c_g)}{\tau}
    \right),
    \label{eq:loss_g}
\end{equation}
where $\sigma$ is the sigmoid function and $\tau$ controls the sharpness of the pairwise preference.

% Neutral Loss
\vspace{2mm}
\noindent\textbf{Neutral calibration loss.}
% \paragraph{Neutral calibration loss.} 
In neutral contexts, no explicit cultural cue points to either entity.
The model should therefore assign similar scores to the local and other-culture entities:
\[
    s_\theta(e_l \mid c_n) \approx s_\theta(e_o \mid c_n).
\]
We define the score gap under model $\theta$ as
\begin{equation}
    \Delta_\theta(c)
    =
    s_\theta(e_l \mid c) - s_\theta(e_o \mid c),
    \label{eq:gap}
\end{equation}
and minimize its squared magnitude in neutral contexts:
\begin{equation}
    \mathcal{L}_n
    =
    \left(\frac{\Delta_\theta(c_n)}{\kappa}\right)^2,
    \label{eq:loss_n}
\end{equation}
where $\kappa$ is a fixed scaling constant.

% Drift Regularization
\vspace{2mm}
\noindent\textbf{Drift regularization.}
% \paragraph{Drift regularization.}
The grounded alignment loss updates LoRA parameters shared across context types, so its gradient steps can also affect entity scores in neutral contexts. The neutral calibration loss alone does not fully stabilize this interaction, and the resulting neutral preferences can diverge from the balanced target. To limit this divergence, we add a regularization term that penalizes deviations of the neutral score gap from its vanilla-model value:
\begin{equation}
    \mathcal{L}_{d}
    =
    \left(
        \frac{
        \Delta_\theta(c_n) - \Delta_{\mathrm{vanilla}}(c_n)
        }{\lambda}
    \right)^2,
    \label{eq:loss_d}
\end{equation}
where $\Delta_{\mathrm{vanilla}}(c_n)$ is the score gap computed with the vanilla model and $\lambda$ controls the strength of the penalty. Together with $\mathcal{L}_n$, this term stabilizes neutral-context behavior during joint optimization with $\mathcal{L}_g$. The two terms are defined against different reference points, zero and the vanilla gap, and the balance between them is set by the weights. Their weighted sum has a single optimum that lies between the two, so together they define one shrinkage target. Appendix~\ref{app:derivation} gives the derivation.

\vspace{2mm}
\noindent\textbf{Goal-aware gradient reconciliation.}
\label{sec:pcgrad}
The grounded and neutral objectives update the same LoRA parameters, so their gradient steps can interfere even though their targets are defined under different contexts. We assign weights $w_{g}$, $w_{n}$, and $w_{d}$ to the three objectives and group them into a grounded direction and a combined neutral-side direction,
\begin{equation}
\begin{aligned}
    \mathbf{v}_g &= \nabla_{\phi}(w_g \mathcal{L}_g), \\
    \mathbf{v}_n &= \nabla_{\phi}(w_n \mathcal{L}_n + w_d \mathcal{L}_d).
\end{aligned}
\end{equation}
The reconciliation described below operates between $\mathbf{v}_{g}$ and $\mathbf{v}_{n}$, not between $\mathcal{L}_{n}$ and $\mathcal{L}_{d}$, which together define the single target described above. We reconcile these directions using Projecting Conflicting Gradients (PCGrad)~\cite{PCGrad}, a gradient-surgery method for multi-objective learning. Our variant adapts PCGrad by weighting the projection with each objective's current distance from its target. When the two gradients conflict ($\mathbf{v}_g \cdot \mathbf{v}_n < 0$), we project each direction against the other with strength proportional to the opposing objective's distance from its target. The projection strength for direction $i$ against direction $j$ is
\begin{equation}
    \beta_{i \leftarrow j} = \frac{d_j}{d_i + d_j},
    \label{eq:pcgrad_beta}
\end{equation}
where $d_i$ and $d_j$ are exponential moving averages of the batch-level CBS distance from the respective targets (0 for grounded, 50 for neutral). The projected gradient is
\begin{equation}
    \tilde{\mathbf{v}}_i = \mathbf{v}_i - \beta_{i \leftarrow j} \frac{\mathbf{v}_i \cdot \mathbf{v}_j}{\|\mathbf{v}_j\|^2} \mathbf{v}_j.
\end{equation}
The objective farther from its target ($d_i > d_j$) receives a smaller $\beta_{i \leftarrow j}$, preserving more of its original update direction. If no conflict exists ($\mathbf{v}_g \cdot \mathbf{v}_n \geq 0$), gradients remain unchanged. The final update direction combines both projected gradients:
\begin{equation}
    \mathbf{v}_f = \tilde{\mathbf{v}}_g + \tilde{\mathbf{v}}_n.
    \label{eq:pcgrad_final}
\end{equation}
We update the LoRA adapters $\phi$ along this direction, completing one training step. This step serves as a safeguard against interference and is not the source of the improvement we report. Replacing it with a plain weighted sum of the same gradients, with all other settings fixed, leaves both $\mathrm{CBS}_{g}$ and $\mathrm{CBS}_{n}$ essentially unchanged (Appendix~\ref{app:gradient}).

\section{Experiments}

% ====== table ======
\begin{table*}[t]
\centering
\setlength{\tabcolsep}{2.2pt}
\renewcommand{\arraystretch}{1.05}
\resizebox{\textwidth}{!}{%
\begin{tabular}{l | ccc c | ccc c | ccc c | ccc c | ccc c | ccc c}
\toprule
\multicolumn{25}{c}{\textbf{Llama-3.1-8B}} \\
\midrule
& \multicolumn{4}{c|}{\textsc{Korean}}
& \multicolumn{4}{c|}{\textsc{Japanese}}
& \multicolumn{4}{c|}{\textsc{Vietnamese}}
& \multicolumn{4}{c|}{\textsc{Hindi}}
& \multicolumn{4}{c|}{\textsc{Arabic}}
& \multicolumn{4}{c}{\textcolor{red}{\textsc{Avg}}} \\
\cmidrule(lr){2-5}\cmidrule(lr){6-9}\cmidrule(lr){10-13}\cmidrule(lr){14-17}\cmidrule(lr){18-21}\cmidrule(lr){22-25}
Method
& \CBSg$\downarrow$ & {\scriptsize$\Delta g$} & \CBSn$\!\to\!50$ & CS$\uparrow$
& \CBSg$\downarrow$ & {\scriptsize$\Delta g$} & \CBSn$\!\to\!50$ & CS$\uparrow$
& \CBSg$\downarrow$ & {\scriptsize$\Delta g$} & \CBSn$\!\to\!50$ & CS$\uparrow$
& \CBSg$\downarrow$ & {\scriptsize$\Delta g$} & \CBSn$\!\to\!50$ & CS$\uparrow$
& \CBSg$\downarrow$ & {\scriptsize$\Delta g$} & \CBSn$\!\to\!50$ & CS$\uparrow$
& \CBSg$\downarrow$ & {\scriptsize$\Delta g$} & \CBSn$\!\to\!50$ & CS$\uparrow$ \\
\midrule
Vanilla
  & \score{47.0}{2.8} & -- & \textbf{\score{49.4}{2.1}} & \underline{2.4}
  & \score{36.1}{2.4} & -- & \score{42.7}{3.6} & 6.6
  & \underline{\score{12.8}{1.8}} & -- & \score{19.7}{1.7} & 6.9
  & \underline{\score{32.2}{3.7}} & -- & \score{41.0}{1.7} & 8.8
  & \underline{\score{37.5}{1.9}} & -- & \score{41.9}{2.8} & 4.4
  & \underline{\score{33.1}{5.6}} & -- & \score{38.9}{4.9} & 5.8 \\
BiasUnlearn
  & \score{47.0}{2.9} & {\scriptsize+0.0} & \underline{\score{49.3}{2.2}} & 2.3
  & \score{40.6}{4.5} & {\scriptsize+4.5} & \underline{\score{48.1}{6.8}} & \underline{7.5}
  & \score{21.6}{3.3} & {\scriptsize+8.8} & \score{30.2}{3.3} & \underline{8.6}
  & \score{36.4}{5.3} & {\scriptsize+4.2} & \underline{\score{47.7}{3.3}} & \underline{11.3}
  & \score{42.3}{3.2} & {\scriptsize+4.8} & \textbf{\score{48.4}{4.5}} & \underline{6.1}
  & \score{37.6}{4.3} & {\scriptsize+4.5} & \underline{\score{44.7}{3.6}} & \underline{7.2} \\
BiasEdit
  & \underline{\score{20.8}{3.8}} & {\scriptsize$-$26.2} & \score{21.8}{3.9} & 1.0
  & \underline{\score{27.6}{2.2}} & {\scriptsize$-$8.5} & \score{28.0}{1.7} & 0.4
  & \score{32.4}{2.7} & {\scriptsize+19.6} & \textbf{\score{36.4}{3.3}} & 4.0
  & \score{46.4}{2.6} & {\scriptsize+14.2} & \score{46.6}{2.6} & 0.2
  & \score{39.2}{2.8} & {\scriptsize+1.7} & \score{38.9}{2.7} & 0.3
  & \score{33.3}{4.4} & {\scriptsize+0.2} & \score{34.3}{4.2} & 1.2 \\
\textbf{CoCoA}
  & \textbf{\score{3.6}{0.5}} & {\scriptsize$-$43.4} & \score{51.3}{1.9} & \textbf{47.7}
  & \textbf{\score{6.3}{1.1}} & {\scriptsize$-$29.8} & \textbf{\score{49.4}{5.6}} & \textbf{43.1}
  & \textbf{\score{7.6}{1.9}} & {\scriptsize$-$5.2} & \underline{\score{36.0}{4.2}} & \textbf{28.4}
  & \textbf{\score{15.4}{3.0}} & {\scriptsize$-$16.8} & \textbf{\score{52.2}{3.7}} & \textbf{36.8}
  & \textbf{\score{12.2}{2.0}} & {\scriptsize$-$25.3} & \underline{\score{52.3}{3.0}} & \textbf{40.1}
  & \textbf{\score{9.0}{2.1}} & {\scriptsize$-$24.1} & \textbf{\score{48.2}{3.0}} & \textbf{39.2} \\

\midrule
\multicolumn{25}{c}{\textbf{Qwen3-8B}} \\
\midrule
& \multicolumn{4}{c|}{\textsc{Korean}}
& \multicolumn{4}{c|}{\textsc{Japanese}}
& \multicolumn{4}{c|}{\textsc{Vietnamese}}
& \multicolumn{4}{c|}{\textsc{Hindi}}
& \multicolumn{4}{c|}{\textsc{Arabic}}
& \multicolumn{4}{c}{\textcolor{red}{\textsc{Avg}}} \\
\cmidrule(lr){2-5}\cmidrule(lr){6-9}\cmidrule(lr){10-13}\cmidrule(lr){14-17}\cmidrule(lr){18-21}\cmidrule(lr){22-25}
Method
& \CBSg$\downarrow$ & {\scriptsize$\Delta g$} & \CBSn$\!\to\!50$ & CS$\uparrow$
& \CBSg$\downarrow$ & {\scriptsize$\Delta g$} & \CBSn$\!\to\!50$ & CS$\uparrow$
& \CBSg$\downarrow$ & {\scriptsize$\Delta g$} & \CBSn$\!\to\!50$ & CS$\uparrow$
& \CBSg$\downarrow$ & {\scriptsize$\Delta g$} & \CBSn$\!\to\!50$ & CS$\uparrow$
& \CBSg$\downarrow$ & {\scriptsize$\Delta g$} & \CBSn$\!\to\!50$ & CS$\uparrow$
& \CBSg$\downarrow$ & {\scriptsize$\Delta g$} & \CBSn$\!\to\!50$ & CS$\uparrow$ \\
\midrule
Vanilla
  & \score{50.5}{4.6} & -- & \score{52.8}{5.2} & 2.3
  & \score{43.7}{3.1} & -- & \underline{\score{44.3}{3.2}} & 0.6
  & \underline{\score{22.2}{2.9}} & -- & \score{28.7}{2.3} & 6.5
  & \underline{\score{42.8}{1.5}} & -- & \underline{\score{48.1}{2.9}} & 5.3
  & \underline{\score{36.4}{3.8}} & -- & \score{38.2}{2.8} & 1.8
  & \score{39.1}{4.7} & -- & \score{42.4}{4.2} & 3.3 \\
BiasUnlearn
  & \score{48.8}{6.0} & {\scriptsize$-$1.7} & \textbf{\score{51.2}{6.5}} & \underline{2.4}
  & \score{45.5}{3.5} & {\scriptsize+1.8} & \textbf{\score{46.1}{3.3}} & 0.6
  & \score{23.8}{3.2} & {\scriptsize+1.6} & \score{30.4}{2.1} & \underline{6.6}
  & \score{43.5}{2.3} & {\scriptsize+0.7} & \textbf{\score{48.9}{3.6}} & \underline{5.4}
  & \score{40.6}{4.1} & {\scriptsize+4.2} & \underline{\score{43.7}{3.2}} & \underline{3.1}
  & \score{40.4}{4.3} & {\scriptsize+1.3} & \underline{\score{44.1}{3.6}} & \underline{3.6} \\
BiasEdit
  & \underline{\score{20.5}{3.1}} & {\scriptsize$-$30.0} & \score{20.1}{3.1} & 0.4
  & \underline{\score{28.3}{3.3}} & {\scriptsize$-$15.4} & \score{29.1}{3.6} & 0.8
  & \score{34.6}{0.7} & {\scriptsize+12.4} & \underline{\score{37.6}{2.1}} & 3.0
  & \score{43.7}{2.9} & {\scriptsize+0.9} & \score{43.4}{3.1} & 0.3
  & \score{36.9}{3.1} & {\scriptsize+0.5} & \score{36.4}{3.3} & 0.5
  & \underline{\score{32.8}{3.9}} & {\scriptsize$-$6.3} & \score{33.3}{3.9} & 1.0 \\
\textbf{CoCoA}
  & \textbf{\score{6.7}{1.6}} & {\scriptsize$-$43.8} & \underline{\score{51.7}{5.7}} & \textbf{45.0}
  & \textbf{\score{18.9}{1.7}} & {\scriptsize$-$24.8} & \score{55.8}{2.9} & \textbf{36.9}
  & \textbf{\score{17.9}{5.1}} & {\scriptsize$-$4.3} & \textbf{\score{43.6}{3.4}} & \textbf{25.7}
  & \textbf{\score{30.5}{2.1}} & {\scriptsize$-$12.3} & \score{54.8}{4.5} & \textbf{24.3}
  & \textbf{\score{14.6}{2.5}} & {\scriptsize$-$21.8} & \textbf{\score{47.3}{3.3}} & \textbf{32.7}
  & \textbf{\score{17.7}{3.8}} & {\scriptsize$-$21.4} & \textbf{\score{50.6}{2.2}} & \textbf{32.9} \\

\midrule
\multicolumn{25}{c}{\textbf{Gemma-3-12B-pt}} \\
\midrule
& \multicolumn{4}{c|}{\textsc{Korean}}
& \multicolumn{4}{c|}{\textsc{Japanese}}
& \multicolumn{4}{c|}{\textsc{Vietnamese}}
& \multicolumn{4}{c|}{\textsc{Hindi}}
& \multicolumn{4}{c|}{\textsc{Arabic}}
& \multicolumn{4}{c}{\textcolor{red}{\textsc{Avg}}} \\
\cmidrule(lr){2-5}\cmidrule(lr){6-9}\cmidrule(lr){10-13}\cmidrule(lr){14-17}\cmidrule(lr){18-21}\cmidrule(lr){22-25}
Method
& \CBSg$\downarrow$ & {\scriptsize$\Delta g$} & \CBSn$\!\to\!50$ & CS$\uparrow$
& \CBSg$\downarrow$ & {\scriptsize$\Delta g$} & \CBSn$\!\to\!50$ & CS$\uparrow$
& \CBSg$\downarrow$ & {\scriptsize$\Delta g$} & \CBSn$\!\to\!50$ & CS$\uparrow$
& \CBSg$\downarrow$ & {\scriptsize$\Delta g$} & \CBSn$\!\to\!50$ & CS$\uparrow$
& \CBSg$\downarrow$ & {\scriptsize$\Delta g$} & \CBSn$\!\to\!50$ & CS$\uparrow$
& \CBSg$\downarrow$ & {\scriptsize$\Delta g$} & \CBSn$\!\to\!50$ & CS$\uparrow$ \\
\midrule
Vanilla
  & \score{50.6}{2.1} & -- & \underline{\score{49.7}{3.2}} & 0.9
  & \score{35.9}{2.0} & -- & \score{36.7}{1.2} & 0.8
  & \score{43.3}{2.8} & -- & \score{41.3}{4.3} & \underline{2.0}
  & \score{53.3}{3.4} & -- & \score{52.7}{3.7} & 0.6
  & \score{43.3}{2.6} & -- & \score{43.2}{3.4} & 0.1
  & \score{45.3}{3.0} & -- & \score{44.7}{2.8} & 0.9 \\
BiasUnlearn
  & \score{50.4}{2.3} & {\scriptsize$-$0.2} & \score{49.5}{3.5} & 0.9
  & \score{36.1}{1.9} & {\scriptsize+0.2} & \underline{\score{37.0}{1.3}} & \underline{0.9}
  & \score{43.2}{2.8} & {\scriptsize$-$0.1} & \score{41.3}{4.4} & 1.9
  & \score{53.1}{3.7} & {\scriptsize$-$0.2} & \score{52.4}{4.0} & 0.7
  & \score{43.8}{2.6} & {\scriptsize+0.5} & \underline{\score{43.8}{3.5}} & 0.0
  & \score{45.3}{2.8} & {\scriptsize+0.0} & \underline{\score{44.8}{2.7}} & \underline{0.9} \\
BiasEdit
  & \underline{\score{49.4}{2.5}} & {\scriptsize$-$1.2} & \score{48.3}{3.4} & \underline{1.1}
  & \underline{\score{35.4}{1.6}} & {\scriptsize$-$0.5} & \score{36.1}{1.4} & 0.7
  & \underline{\score{43.0}{2.6}} & {\scriptsize$-$0.3} & \underline{\score{41.4}{4.3}} & 1.6
  & \underline{\score{52.6}{3.3}} & {\scriptsize$-$0.7} & \textbf{\score{51.7}{3.1}} & \underline{0.9}
  & \underline{\score{42.5}{2.2}} & {\scriptsize$-$0.8} & \score{42.6}{3.5} & \underline{0.1}
  & \underline{\score{44.6}{2.9}} & {\scriptsize$-$0.7} & \score{44.0}{2.7} & 0.9 \\
\textbf{CoCoA}
  & \textbf{\score{25.7}{2.7}} & {\scriptsize$-$24.9} & \textbf{\score{50.3}{3.1}} & \textbf{24.6}
  & \textbf{\score{19.9}{1.0}} & {\scriptsize$-$16.0} & \textbf{\score{38.7}{1.6}} & \textbf{18.8}
  & \textbf{\score{37.1}{6.6}} & {\scriptsize$-$6.2} & \textbf{\score{42.4}{4.5}} & \textbf{5.3}
  & \textbf{\score{42.3}{2.7}} & {\scriptsize$-$11.0} & \underline{\score{52.2}{3.1}} & \textbf{9.9}
  & \textbf{\score{33.5}{3.7}} & {\scriptsize$-$9.8} & \textbf{\score{44.3}{3.5}} & \textbf{10.8}
  & \textbf{\score{31.7}{3.9}} & {\scriptsize$-$13.6} & \textbf{\score{45.6}{2.5}} & \textbf{13.9} \\

\midrule
\multicolumn{25}{c}{\textbf{Mistral-7B-v0.3}} \\
\midrule
& \multicolumn{4}{c|}{\textsc{Korean}}
& \multicolumn{4}{c|}{\textsc{Japanese}}
& \multicolumn{4}{c|}{\textsc{Vietnamese}}
& \multicolumn{4}{c|}{\textsc{Hindi}}
& \multicolumn{4}{c|}{\textsc{Arabic}}
& \multicolumn{4}{c}{\textcolor{red}{\textsc{Avg}}} \\
\cmidrule(lr){2-5}\cmidrule(lr){6-9}\cmidrule(lr){10-13}\cmidrule(lr){14-17}\cmidrule(lr){18-21}\cmidrule(lr){22-25}
Method
& \CBSg$\downarrow$ & {\scriptsize$\Delta g$} & \CBSn$\!\to\!50$ & CS$\uparrow$
& \CBSg$\downarrow$ & {\scriptsize$\Delta g$} & \CBSn$\!\to\!50$ & CS$\uparrow$
& \CBSg$\downarrow$ & {\scriptsize$\Delta g$} & \CBSn$\!\to\!50$ & CS$\uparrow$
& \CBSg$\downarrow$ & {\scriptsize$\Delta g$} & \CBSn$\!\to\!50$ & CS$\uparrow$
& \CBSg$\downarrow$ & {\scriptsize$\Delta g$} & \CBSn$\!\to\!50$ & CS$\uparrow$
& \CBSg$\downarrow$ & {\scriptsize$\Delta g$} & \CBSn$\!\to\!50$ & CS$\uparrow$ \\
\midrule
Vanilla
  & \score{52.3}{2.9} & -- & \underline{\score{57.8}{2.9}} & \underline{5.5}
  & \score{31.9}{2.1} & -- & \score{36.2}{1.4} & 4.3
  & \underline{\score{37.6}{2.3}} & -- & \score{43.9}{3.1} & 6.3
  & \score{49.1}{2.8} & -- & \score{52.3}{2.7} & \underline{3.2}
  & \score{42.1}{0.8} & -- & \score{39.8}{2.5} & \underline{2.3}
  & \score{42.6}{3.8} & -- & \score{46.0}{3.9} & \underline{4.3} \\
BiasUnlearn
  & \score{46.4}{3.7} & {\scriptsize$-$5.9} & \textbf{\score{50.8}{5.7}} & 4.4
  & \score{39.5}{4.3} & {\scriptsize+7.6} & \underline{\score{44.8}{5.8}} & \underline{5.3}
  & \score{40.7}{4.2} & {\scriptsize+3.1} & \score{47.4}{5.9} & \underline{6.7}
  & \underline{\score{47.6}{3.2}} & {\scriptsize$-$1.5} & \textbf{\score{50.3}{3.7}} & 2.7
  & \score{48.7}{2.4} & {\scriptsize+6.6} & \textbf{\score{47.3}{3.2}} & 1.4
  & \score{44.6}{1.8} & {\scriptsize+2.0} & \textbf{\score{48.1}{1.1}} & 4.1 \\
BiasEdit
  & \underline{\score{24.7}{2.2}} & {\scriptsize$-$27.6} & \score{25.6}{1.4} & 0.9
  & \underline{\score{28.7}{3.3}} & {\scriptsize$-$3.2} & \score{29.3}{4.2} & 0.6
  & \score{46.8}{1.9} & {\scriptsize+9.2} & \underline{\score{51.0}{1.8}} & 4.2
  & \score{47.8}{2.0} & {\scriptsize$-$1.3} & \score{46.7}{2.0} & 1.1
  & \underline{\score{38.9}{2.3}} & {\scriptsize$-$3.2} & \score{37.9}{2.4} & 1.0
  & \underline{\score{37.4}{4.7}} & {\scriptsize$-$5.2} & \score{38.1}{4.8} & 1.6 \\
\textbf{CoCoA}
  & \textbf{\score{3.1}{0.6}} & {\scriptsize$-$49.2} & \score{58.6}{3.0} & \textbf{55.5}
  & \textbf{\score{8.1}{1.4}} & {\scriptsize$-$23.8} & \textbf{\score{49.8}{3.6}} & \textbf{41.7}
  & \textbf{\score{10.2}{3.9}} & {\scriptsize$-$27.4} & \textbf{\score{50.9}{5.0}} & \textbf{40.7}
  & \textbf{\score{21.2}{2.3}} & {\scriptsize$-$27.9} & \underline{\score{50.7}{1.7}} & \textbf{29.5}
  & \textbf{\score{7.8}{1.8}} & {\scriptsize$-$34.3} & \underline{\score{56.2}{2.4}} & \textbf{48.4}
  & \textbf{\score{10.1}{3.0}} & {\scriptsize$-$32.5} & \underline{\score{53.2}{1.8}} & \textbf{43.2} \\
\bottomrule
\end{tabular}}
\caption{Cultural Bias Scores on five language settings across four backbone models (5-fold CV, mean$\pm$SE). CBS$_g$ measures Western preference in grounded contexts, CBS$_n$ measures balance in neutral contexts, and CS $= |$CBS$_n - $CBS$_g|$ summarizes context sensitivity. $\Delta g$ is the change in CBS$_g$ relative to the vanilla model. \textbf{Bold} marks the best value per column, \underline{underline} the second best. 
The remaining five language settings are in Table~\ref{tab:main_rest}.
}
\label{tab:main}
\end{table*}

% \dongho{Can we reduce the number of languages shown in this table for better readability and move the remaining results to the appendix? Also, we could use $\downarrow$ and $\uparrow$ indicators in the table as well. + I think we should emphasize CoCoA consistently achieves large reductions in CBS$_g$ while preserVietnameseng CBS$_n$ near the desired 50\% target, whereas prior methods either fail to substantially reduce grounded bias or collapse neutral calibration. To show this, we may need to change the table structure. Rather than showing CBS$_g$ and CBS$_n$ raw score itself in the table, you can also show $\Delta$CBS$_g$, $\Delta$CBS$_n$ to show large reduction in CBS$_g$ while maintaining CBS$_n$}

\subsection{Datasets and Experimental Setup}
\label{sec:setup}
\noindent\textbf{CAMeL and Camellia.}
We train and evaluate CoCoA on CAMeL~\cite{camel} and Camellia~\cite{camellia}, two entity-centric cultural bias benchmarks that pair culturally associated entities with naturally occurring contexts. Both benchmarks share the same evaluation structure, pairing local entities (Arab in CAMeL, Asian in Camellia) against Western counterparts across grounded and neutral contexts. They differ in scope, with CAMeL focusing on Arabic and Camellia covering nine Asian languages, but together they span seven cultural groups across ten language settings. In both benchmarks, native speakers of the respective language manually assign the grounded and neutral labels. A context is labeled grounded when annotators judge target-culture entities to be uniquely appropriate in it, and neutral otherwise.

\vspace{2mm}
\noindent\textbf{Backbone models.}
We evaluate CoCoA on four language models: Llama-3.1-8B~\cite{llama3}, Qwen3-8B~\cite{qwen3}, Mistral-7B-v0.3~\cite{mistral}, and Gemma-3-12B-pt~\cite{gemma3}. While we primarily focus on pretrained models to isolate raw cultural preferences from post-training confounds, we include Qwen3-8B to verify whether CoCoA generalizes to instruction-tuned backbones.

\vspace{2mm}
\noindent\textbf{Evaluation protocol.}
All experiments use 5-fold cross-validation. Within each language setting we partition the contexts and the entity sets into five folds before any pairing is constructed, so no context, entity, or entity pair is shared across splits, and evaluation uses entities the model has not been trained on. In each fold, three splits go to training, one to validation, and one to evaluation. We compute CBS on the evaluation fold using pairwise comparisons between local and other-culture entities, up to $30 \times 30$ comparisons per context. All methods use PMI normalization with $\alpha{=}1.0$ for evaluation. Each reported number reflects the mean and standard error across the five folds.

\subsection{Baselines}
\label{sec:baselines}
We compare CoCoA against BiasEdit~\cite{biasedit} and BiasUnlearn~\cite{biasunlearn}, two social bias mitigation methods originally designed for stereotype debiasing. BiasEdit edits MLP weight matrices through a meta-learned hypernetwork, while BiasUnlearn applies gradient-based unlearning with forget and retain objectives. Since we are not aware of prior methods that target entity-centric cultural bias directly, we adapt both to our setting by replacing their social group pairs with local--other-culture entity pairs and training on culturally neutral contexts, where balanced preferences are desired. Both baselines share the same data splits and evaluation protocol as CoCoA. Detailed adaptation procedures are provided in Appendix~\ref{app:baselines}. We also compare against two culture-aware prompting baselines from \citet{culturespa}, which \citet{culturesteer} adopt as cultural bias baselines under the names Culture-Specific Prompt (CSP) and Cross-Culture Think (CCT). Both condition the model at inference time without updating parameters and are reported in Appendix~\ref{app:prompting}.

\subsection{Main Results}
\label{sec:results}
Table~\ref{tab:main} reports CBS$_g$ and CBS$_n$ on five representative language settings, and Table~\ref{tab:main_rest} reports the remaining five. Lower CBS$_g$ indicates a stronger preference for local entities in grounded contexts, while CBS$_n$ closer to 50 indicates balanced behavior in neutral contexts.

\vspace{2mm}
\noindent\textbf{CoCoA improves grounded-context preference while preserving neutral calibration.}
Across all four backbone families, CoCoA consistently reduces CBS$_g$ compared to the vanilla model. The average CBS$_g$ drops from 33.1 to 9.0 for Llama-3.1-8B, from 39.1 to 17.7 for Qwen3-8B, from 45.3 to 31.7 for Gemma-3-12B-pt, and from 42.6 to 10.1 for Mistral-7B-v0.3. At the same time, CBS$_n$ remains close to the balanced target for Llama-3.1-8B and Qwen3-8B, with average values of 48.2 and 50.6. Mistral-7B-v0.3 also stays close at 53.2, while Gemma-3-12B-pt shows a larger gap at 45.6. The appendix results follow the same overall trend on the remaining language settings. Averaged over all ten language settings and the four backbones, $\mathrm{CBS}_{g}$ falls from 43 to 24 while $\mathrm{CBS}_{n}$ reaches 50.2.

\vspace{2mm}
\noindent\textbf{Neutralization baselines do not produce context-conditional behavior.}
BiasUnlearn preserves neutral-context behavior better than BiasEdit, but does not meaningfully reduce CBS$_g$. For Llama-3.1-8B, the average CBS$_g$ moves from 33.1 to 37.6, while CBS$_n$ reaches 44.7. BiasEdit reduces CBS$_g$ in some language settings but increases it in others, and the same correction often distorts neutral-context behavior. For example, on Korean with Llama-3.1-8B, BiasEdit reduces CBS$_g$ from 47.0 to 20.8 but also shifts CBS$_n$ from 49.4 to 21.8. These patterns suggest that neutralization-focused adaptations alone do not produce the desired context-conditional behavior.

\vspace{2mm}
\noindent\textbf{The context-sensitivity gap captures the same pattern.}
We also report the context-sensitivity gap, CS $= |$CBS$_n -$ CBS$_g|$, as an auxiliary summary of how differently a model behaves in grounded and neutral contexts. This gap should be read together with CBS$_n$, since a large gap alone is not desirable if neutral-context behavior moves far from 50. Under this reading, CoCoA shows the clearest context-conditional pattern by reducing CBS$_g$ while keeping CBS$_n$ close to 50. This holds most strongly for Llama-3.1-8B, Qwen3-8B, and Mistral-7B-v0.3, while Gemma-3-12B-pt follows the same trend with smaller separation. The neutralization baselines show weaker separation between the two context types, and when separation does appear, it often comes from neutral-context behavior drifting away from 50 rather than from improved grounded-context preference.
\section{Analysis}
\label{sec:analysis}

\begin{table}[t]
\centering
\footnotesize
\setlength{\tabcolsep}{4.5pt}
\renewcommand{\arraystretch}{1.05}
\begin{tabular}{@{}llcc@{}}
\toprule
Objective & Weight & CBS$_g\!\downarrow$ & CBS$_n\!\to\!50$ \\
\midrule
\multirow{5}{*}{Grounded}
    & $w_g=0$   & $54.6{\pm}1.3$ & $60.7{\pm}1.4$ \\
    & $w_g=0.5$ & $15.8{\pm}1.4$ & $53.6{\pm}1.4$ \\
    & $w_g=1.0$ & $15.0{\pm}1.3$ & $53.7{\pm}1.4$ \\
    & $w_g=1.5$ & $14.7{\pm}1.3$ & $52.8{\pm}1.4$ \\
    & $w_g=2.0$ & $14.5{\pm}1.3$ & $52.9{\pm}1.4$ \\
\midrule
\multirow{5}{*}{Neutral}
    & $w_n=0$   & $13.4{\pm}1.1$ & $37.0{\pm}1.1$ \\
    & $w_n=1.0$ & $14.3{\pm}1.2$ & $49.0{\pm}1.3$ \\
    & $w_n=2.0$ & $15.0{\pm}1.3$ & $53.2{\pm}1.4$ \\
    & $w_n=3.0$ & $15.3{\pm}1.3$ & $55.3{\pm}1.4$ \\
    & $w_n=4.0$ & $16.0{\pm}1.3$ & $56.4{\pm}1.6$ \\
\midrule
\multirow{5}{*}{Drift}
    & $w_d=0$   & $15.2{\pm}1.3$ & $58.8{\pm}1.5$ \\
    & $w_d=0.5$ & $15.1{\pm}1.3$ & $55.9{\pm}1.5$ \\
    & $w_d=1.0$ & $15.1{\pm}1.3$ & $53.3{\pm}1.4$ \\
    & $w_d=2.0$ & $15.1{\pm}1.3$ & $49.8{\pm}1.3$ \\
    & $w_d=3.0$ & $15.3{\pm}1.3$ & $47.1{\pm}1.3$ \\
\bottomrule
\end{tabular}
    \caption{Objective weight sensitivity on Llama-3.1-8B, averaged across 10 language settings (5-fold CV). Each block is an independent sweep of one weight with the others fixed at $w_g{=}1.0$, $w_n{=}2.0$, $w_d{=}1.0$.}
\label{tab:weight_sensitivity}
\end{table}

\noindent\textbf{Objective ablation and weight sensitivity.} To examine the role of each objective, we ablate one at a time by setting its weight to zero, and sweep nonzero values to assess sensitivity. Table~\ref{tab:weight_sensitivity} reports these sweeps.
The grounded loss directly controls local preference in grounded contexts. Removing it ($w_g{=}0$) raises CBS$_g$ to 54.6, confirming that the model relies on $\mathcal{L}_g$ for grounded-context preference. Increasing $w_g$ beyond 0.5 yields diminishing returns on CBS$_g$ while leaving CBS$_n$ largely unaffected, suggesting that $\mathcal{L}_g$ mainly supplies the grounded preference signal.
The neutral loss controls whether grounded-context preference also spills into neutral contexts. Without it ($w_n{=}0$), CBS$_g$ stays low at 13.4 but CBS$_n$ drops to 37.0, indicating overcorrection toward local entities in neutral contexts. Increasing $w_n$ moves CBS$_n$ back toward the balanced region, though larger values push it above 50, introducing a calibration trade-off.
Drift regularization plays a stabilizing role. Without it ($w_d{=}0$), CBS$_n$ shifts to 58.8 while CBS$_g$ stays at 15.2. Increasing $w_d$ steadily brings CBS$_n$ back toward 50 without affecting CBS$_g$, confirming that $\mathcal{L}_d$ stabilizes neutral-context behavior during joint optimization without interfering with grounded-context preference.

% per-pair gap
\begin{figure}[t]
    \centering
    \includegraphics[width=\columnwidth]{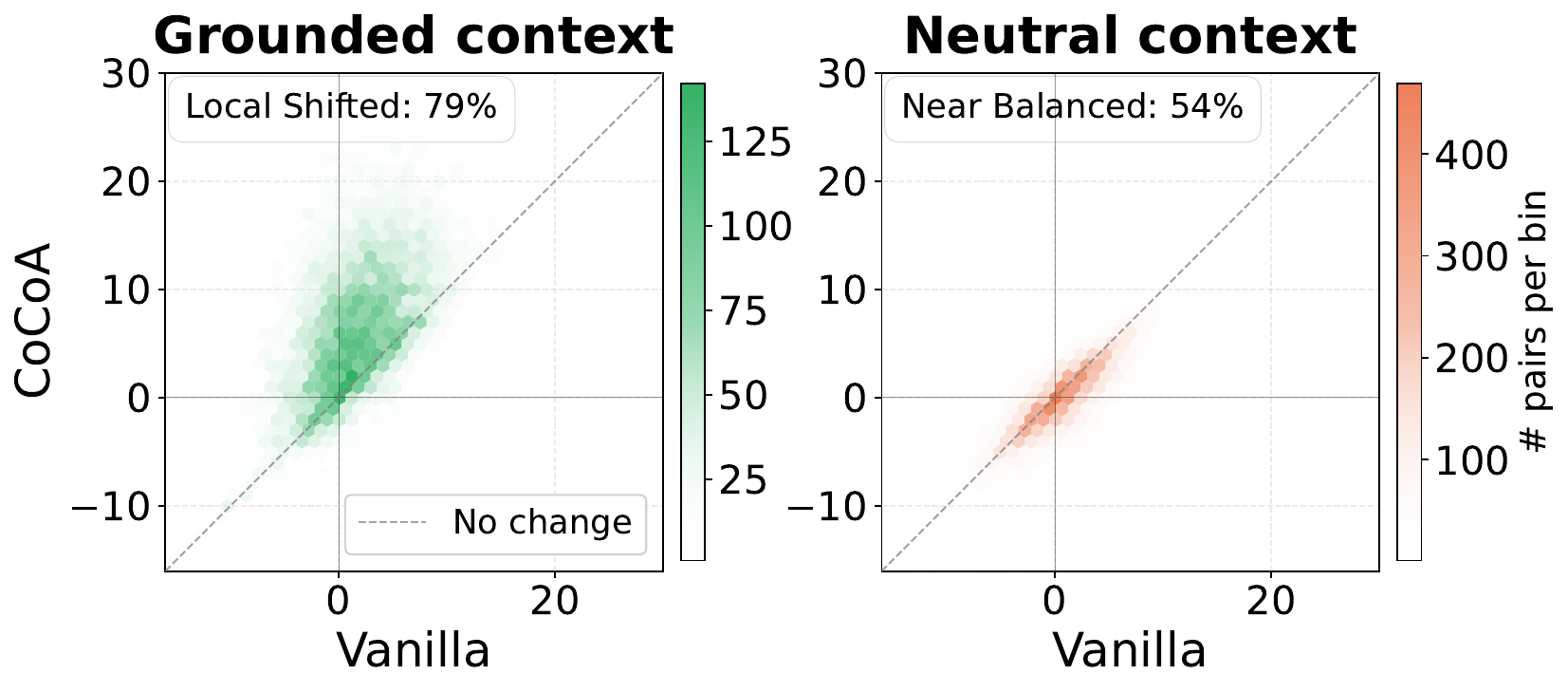}
    \caption{Per-pair score gap of local--Western entities under the vanilla model (x-axis) vs.\ CoCoA (y-axis) on Llama-3.1-8B. The diagonal marks no change.}
    \label{fig:per_pair}
\end{figure}

% Per-Entity Pair
\vspace{2mm}
\noindent\textbf{Per-pair preference shift.} 
To examine whether CoCoA changes entity preferences differently across context types, we compare the local--Western score gap under the vanilla model and under CoCoA. For each entity pair and context, we compute the gap $s(e_l \mid c) - s(e_o \mid c)$. Figure~\ref{fig:per_pair} plots the vanilla gap on the x-axis and the CoCoA gap on the y-axis. Points above the dashed diagonal indicate that CoCoA increases local preference relative to the vanilla model, while points near the origin indicate balanced preferences.
In grounded contexts, 79\% of pairs shift above the diagonal toward stronger local preference, confirming that CoCoA amplifies the relative preference for the entity associated with the cultural cue. In neutral contexts, the distribution clusters near both the origin and the diagonal, indicating that CoCoA maintains balanced preferences without substantially altering the vanilla model's neutral behavior. Among all neutral-context pairs, 54\% show a reduced absolute gap after CoCoA, suggesting that the framework moves preferences toward balance rather than away from it. Appendix~\ref{app:decomposition} decomposes these shifts into the local and Western sides.

\begin{figure}[t]
    \centering
    \includegraphics[width=\columnwidth]{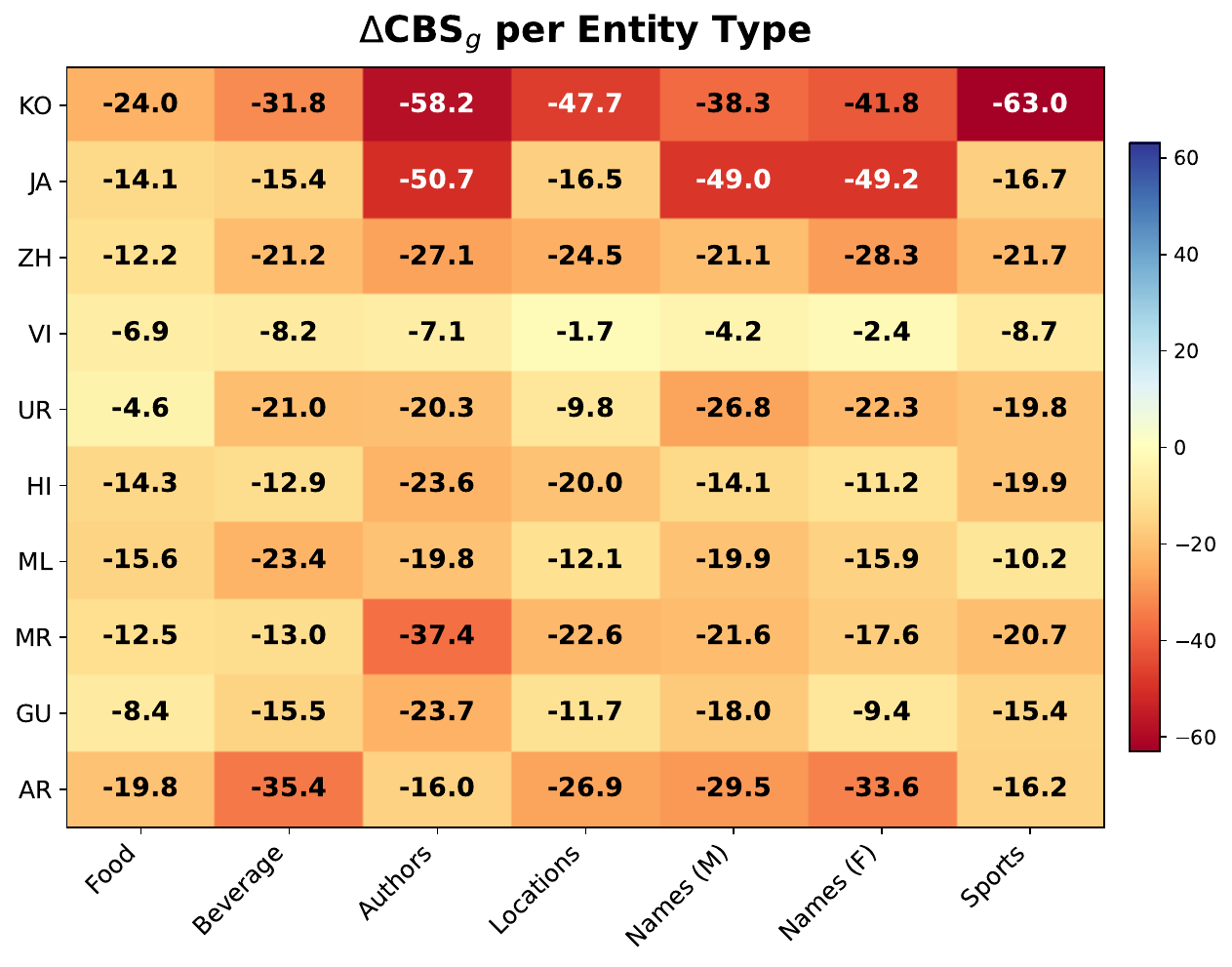}
    \caption{Per-entity-type $\Delta\mathrm{CBS}_g$ on Llama-3.1-8B. Negative values indicate reduced Western preference in grounded contexts.}
    \label{fig:entity_type}
\end{figure}

\vspace{2mm}
\noindent\textbf{Per-entity-type breakdown.} 
To examine whether grounded-context improvements are concentrated in a few entity types or distributed more broadly, we compute $\Delta\mathrm{CBS}_g = \mathrm{CBS}_g^{\text{CoCoA}} - \mathrm{CBS}_g^{\text{vanilla}}$ for each language setting and entity type. Negative values indicate that CoCoA reduces Western preference in grounded contexts.
Figure~\ref{fig:entity_type} shows negative $\Delta\mathrm{CBS}_g$ values across all language--entity type combinations, confirming that improvements are not confined to a subset of entity types. The reductions are largest for Korean, Japanese, and Arabic, while Vietnamese shows smaller but still consistent improvements. Within each language setting, the magnitude varies across entity types, with some categories such as Authors and Names showing particularly large shifts.

% per-entity type

% \subsection{Further Analysis}
% \label{sec:impact}

\vspace{2mm}
\noindent\textbf{General modeling performance is preserved.}
To check whether CoCoA affects general model performance, we evaluate the CoCoA-trained Llama-3.1-8B checkpoints on five standard benchmarks: MMLU~\cite{mmlu} (multitask language understanding), HellaSwag~\cite{hellaswag} (commonsense completion), ARC-Challenge~\cite{arc} (science reasoning), WinoGrande~\cite{winogrande} (pronoun resolution), and TruthfulQA~\cite{tqa} (truthful question answering). We report accuracy for MMLU and WinoGrande, length-normalized accuracy for HellaSwag and ARC-Challenge, and MC2 for TruthfulQA, following the standard evaluation setup for each benchmark. For CoCoA, we average performance across all 50 language-setting--fold checkpoints and compare it with the vanilla Llama-3.1-8B model.

\begin{table}[!b]
\centering
\small
\setlength{\tabcolsep}{3.5pt}
\begin{tabular}{lccccc}
\toprule
\multirow{2}{*}[-0.1ex]{Method} 
       & \textbf{MMLU} & \textbf{HSwag} & \textbf{ARC-C} & \textbf{WGrande} & \textbf{TQA} \\
       & \scriptsize(5-shot) & \scriptsize(10-shot) & \scriptsize(25-shot) & \scriptsize(5-shot) & \scriptsize(0-shot) \\
\midrule
Vanilla & 65.34 & 82.24 & 58.19 & 78.37 & 44.16 \\
CoCoA   & 65.22 & 82.04 & 57.81 & 78.06 & 44.14 \\
\quad$\Delta$ 
        & \textcolor{gray}{\footnotesize$-$0.12} 
        & \textcolor{gray}{\footnotesize$-$0.20} 
        & \textcolor{gray}{\footnotesize$-$0.38} 
        & \textcolor{gray}{\footnotesize$-$0.31} 
        & \textcolor{gray}{\footnotesize$-$0.02} \\
\bottomrule
\end{tabular}
\caption{General benchmark performance on Llama-3.1-8B. CoCoA scores are averaged across all 50 checkpoints (10 language settings $\times$ 5 folds).}
\label{tab:general_perf}
\end{table}

Table~\ref{tab:general_perf} shows that CoCoA changes benchmark performance only marginally. The largest drop is 0.38 percentage points on ARC-Challenge, and the remaining changes are below 0.32 percentage points. The LoRA-based updates used for cultural alignment preserve general benchmark performance while substantially changing context-sensitive cultural entity preferences.

\section{Conclusion and Future Work}
\label{sec:conclusion}
In this work, we argued that entity-centric cultural bias mitigation requires a context-sensitive approach that differs from conventional social debiasing. While social bias mitigation aims to neutralize group preferences uniformly, cultural associations can be appropriate in context. To address this, we proposed CoCoA, a framework that induces context-conditional cultural preferences through dual-context training, complementary loss functions, and goal-aware gradient optimization. 

Experiments across seven cultural groups, ten language settings, and four backbone models show that CoCoA reduces Western preference in grounded contexts while preserving balance in neutral contexts. Compared with adapted social bias mitigation baselines, CoCoA better distinguishes contexts where a contextually supported entity association is appropriate from contexts where balanced behavior is desired. Because contexts, entities, and pairings are disjoint across splits, this behavior generalizes to unseen instances within the benchmark distribution, though not necessarily to implicit or graded cues in unrestricted text. 

These results suggest that cultural alignment requires objectives sensitive to when cultural preferences are appropriate, not just whether they exist. A natural extension would take CoCoA beyond the current binary local--other-culture pairing to a multicultural setting, where multiple local cultures are aligned simultaneously. This would enable finer-grained control and help prevent preference imbalances across culturally proximate groups, but it requires both multi-way objectives and datasets that capture intercultural distinctions.
Bias reduction also varies across entity types under the current type-agnostic setup (Figure~\ref{fig:entity_type}), and type-level loss reweighting could yield more balanced improvements.

% \clearpage
\section*{Limitations}
While CoCoA performs well on average, its weight setting does not transfer uniformly across configurations. Seven of the forty model--setting configurations use non-default weights. Improvement magnitude also differs across settings in a way the vanilla model's bias level does not explain. On Llama-3.1-8B the Korean, Japanese, and Chinese settings fall by 31.8 points against 17.1 for the five South Asian settings, from vanilla scores of 39.8 and 38.8. Developing culture-adaptive training strategies that reduce manual tuning would improve practical deployability.

Our formulation treats the presence of a cultural cue as binary and does not handle cues that are implicit or graded in strength. Both benchmarks also pair local entities against Western counterparts, so we do not establish how CoCoA behaves when multiple non-Western cultures are aligned at once. Our evaluation focuses on entity-centric fill-in-the-blank tasks and does not cover open-ended generation or downstream cultural reasoning.
\section*{Ethics Statement}
This work aims to improve cultural fairness in language models by making entity preferences more sensitive to context. \method{} encourages culturally appropriate preferences when explicit cultural cues are present, while preserving balanced preferences in culturally neutral contexts. Our method is not intended to erase genuine cultural associations or suppress legitimate cultural expressions.
We use only publicly available models and benchmarks, following their data and model usage policies. Any culturally biased or sensitive examples presented in this paper are for illustrative and evaluation purposes and do not represent the views of the authors. We acknowledge that bias mitigation methods can be misused to steer models toward specific cultural viewpoints and strongly encourage responsible use of \method{}, with careful consideration of the relevant cultural contexts and the communities affected.
% \section{Dataset}

% \clearpage
% Entries for the entire Anthology, followed by custom entries

\section*{Acknowledgments}
We would like to thank the anonymous reviewers for their helpful questions and comments.
This work was partly supported by Institute of Information \& communications Technology Planning \& Evaluation(IITP) grant funded by the Korea government(MSIT)
(IITP-2026-RS-2020-II201821, ICT Creative Consilience Program \& 
RS-2024-00509258 and No. RS-2024-00469482, Global AI Frontier Lab \&
RS-2025-25442569, AI Star Fellowship Support Program(Sungkyunkwan University))
 and the Ministry of Education of the Republic of Korea and the National Research Foundation of Korea (NRF-RS-2025-00523385).

\bibliography{main}

\clearpage
\appendix

\section{Experimental Details}
\label{app:exp_details}

% Implementation Details %
\subsection{Implementation Details}
\label{app:impl}

CoCoA applies LoRA adapters to the attention projection matrices ($W_Q, W_K, W_V, W_O$) with rank $r{=}16$ and scaling factor $\alpha_{\mathrm{LoRA}}{=}32$. Training proceeds for 15 epochs per language using AdamW with learning rate $1 \times 10^{-5}$ and a cosine schedule with 10\% warmup. Each setting is trained independently. Entity pairs are constructed using an $N{\times}N$ pairing strategy, where equal-sized subsets of local and other-culture entities are sampled and all combinations are formed, yielding diverse pairings across epochs. Default loss weights are $w_g{=}1.0$, $w_n{=}2.0$, and $w_d{=}1.0$, with contrastive temperature $\tau{=}1.0$. All experiments run on NVIDIA A100 80GB GPUs. The full set of default hyperparameters is listed in Table~\ref{tab:cocoa-hparams}, and Appendix~\ref{app:weights} describes the seven configurations that depart from the default weights.

\subsection{Objective Weights}
\label{app:weights}

\begin{table}[b!]
\centering
\small
\setlength{\tabcolsep}{4pt}
\begin{tabular}{llcc}
\toprule
Model & Language & $w_n$ & $w_d$ \\
\midrule
\multirow{3}{*}{Llama}
 & KO & 2.0 & 20.0 \\
 & VI & 2.0 & 0.5 \\
 & HI, MR & 2.0 & 3.0 \\
\midrule
Qwen & KO & 1.0 & 3.0 \\
\midrule
\multirow{2}{*}{Mistral}
 & KO & 1.0 & 30.0 \\
 & HI & 2.0 & 5.0 \\
\bottomrule
\end{tabular}
\caption{Non-default hyperparameters. All unlisted pairs use $w_g{=}1.0$, $w_n{=}2.0$, $w_d{=}1.0$. Gemma uses defaults for all languages.}
\label{tab:hparams}
\end{table}

CoCoA uses a single default weight setting, $w_g{=}1.0$, $w_n{=}2.0$ and $w_d{=}1.0$, listed with the other defaults in Table~\ref{tab:cocoa-hparams}. These values were fixed in advance as a reasonable operating point and are not the outcome of a search.
 
We apply that setting to all 40 model--setting configurations and change it only where the default leaves $\mathrm{CBS}_{n}$ far from 50 on the validation split. Thirty-three configurations keep the default. Of the remaining seven, five change $w_d$ alone and two also change $w_n$. Table~\ref{tab:hparams} lists them.
 
The adjustment is directional. When the default leaves $\mathrm{CBS}_{n}$ above 50, neutral-context preference has drifted toward other-culture entities, and raising $w_d$ holds the neutral gap closer to its vanilla value. When $\mathrm{CBS}_{n}$ falls below 50, $w_d$ is lowered.
Table~\ref{tab:weight_selection} shows the ten Llama-3.1-8B settings. The four that were adjusted are the four whose default $\mathrm{CBS}_{n}$ is farthest from 50, by 10.5 points or more, while the six that keep the default deviate by at most 3.6 points. The six unadjusted settings are identical in the two columns because nothing about their training changed. Averaged over the ten settings, $\mathrm{CBS}_{n}$ moves from 53.5 under the default weights to 50.3 under the selected ones.

\begin{table}[t]
\centering
\small
\begin{tabular}{lccc}
\toprule
 & & \multicolumn{2}{c}{$\mathrm{CBS}_{n}$} \\
\cmidrule(lr){3-4}
Language & $w_d$ & Default & Selected \\
\midrule
KO & 20.0 & 69.1 & 51.3 \\
JA &  1.0 & 49.4 & 49.4 \\
ZH &  1.0 & 49.3 & 49.3 \\
VI &  0.5 & 33.8 & 36.0 \\
UR &  1.0 & 52.4 & 52.4 \\
HI &  3.0 & 62.9 & 52.2 \\
ML &  1.0 & 52.0 & 52.0 \\
MR &  3.0 & 60.5 & 54.7 \\
GU &  1.0 & 53.6 & 53.6 \\
AR &  1.0 & 52.3 & 52.3 \\
\midrule
Avg & & \score{53.5}{1.4} & \score{50.3}{0.9} \\
\bottomrule
\end{tabular}
\caption{Neutral-context behavior under the default weights and under the selected weights, on Llama-3.1-8B over 5-fold cross-validation. Per-setting entries are fold means. Settings with $w_d{=}1.0$ keep the default and are unchanged between the two columns.}
\label{tab:weight_selection}
\end{table}
 
% Baselines %
\subsection{Baseline Adaptation}
\label{app:baselines}

We adapt two social bias mitigation methods to the cultural entity-pair setting, preserving their original training objectives while changing only the data format.

\vspace{2mm}
\noindent\textbf{BiasEdit.}
BiasEdit~\cite{biasedit} uses MALMEN, a meta-learning approach that trains a hypernetwork to produce weight edits for targeted MLP layers. We replace its social group pairs with our local--other-culture entity pairs and train on neutral contexts. The hypernetwork learns to edit MLP representations so that local and other-culture entities receive similar scores in neutral contexts.

\vspace{2mm}
\noindent\textbf{BiasUnlearn.}
BiasUnlearn~\cite{biasunlearn} applies gradient-based unlearning through a dual-pathway mechanism that coordinates forgetting of biased associations with retention of general language ability. We construct forget sets from neutral-context entity pairs where the model shows strong other-culture preference, and retain sets from balanced pairs. Training uses neutral contexts only.

\vspace{2mm}
\noindent\textbf{Fair comparison protocol.}
Both baselines share the same 5-fold data splits, entity prior computation, and PMI-based evaluation protocol as CoCoA. PMI normalization is applied only at evaluation for both baselines, since their token-level training losses are structurally incompatible with entity-level PMI normalization during training.

% ============================================

\section{Additional Results}
\label{app:main_add}
\subsection{Remaining Main Results}
\label{app:remaining}
Table~\ref{tab:main_rest} reports the results of five language settings not included in Table~\ref{tab:main}.

\subsection{Culture-Aware Prompting Baselines}
\label{app:prompting}

CSP and CCT condition the model on cultural information in the prompt and leave its parameters untouched. They test whether the context-conditional behavior CoCoA is trained for can instead be requested at inference time.
 
Both methods come from \citet{culturespa}. Culture-Specific Prompt (CSP) assigns a cultural persona in the prompt, and Cross-Culture Think (CCT) adds an explicit cross-cultural comparison step before the model responds. \citet{culturesteer} adopt both as prompting baselines for cultural bias, and we follow their CSP and CCT naming. We keep both templates and change the task description from word association to sentence completion, so the prompt matches the fill-in-the-blank setting of CAMeL and Camellia. The comparison group in CCT is set to Western cultures, since the entity pairs in both benchmarks contrast a local entity with a Western counterpart. The persona names follow the seven cultural groups of Section~\ref{sec:setup} and not the ten language settings, so the four Indic settings all take the Indian persona.

Table~\ref{tab:prompting_base} reports both methods against the vanilla model
and CoCoA. Neither produces the behavior CoCoA is trained for. On Llama-3.1-8B, $\mathrm{CBS}_{g}$ stays at 36.2 for CSP and 36.9 for CCT against 36.4 for the vanilla model, so asking the model to attend to culture does not make it prefer culturally appropriate entities when a cue is present. The same holds on Qwen3-8B, where CSP reaches 41.1 and CCT 41.7 against a vanilla value of 41.7. CoCoA reaches 15.2 and 24.9 on the two backbones.
 
Neutral contexts move in the wrong direction. Both methods pull $\mathrm{CBS}_{n}$ further below 50 on both backbones. On Llama-3.1-8B it falls from 43.4 to 40.5 under CSP and 42.1 under CCT, and on Qwen3-8B from 45.8 to 43.3 and 43.4. Cultural conditioning is applied to every context, including those with no cue, so it pushes preference toward local entities where the benchmarks intend neither entity to be preferred. CoCoA reaches 50.3 and 50.7. A single prompt cannot hold the two objectives apart.
 
The two approaches also differ in cost, in ways that do not favor either uniformly. CSP and CCT need no training, but the prefixes in Table~\ref{tab:prompt-templates} run to 24 and 39 words and are prepended to every query. CoCoA pays once during training and adds nothing at inference.
 
\begin{table}[t]
\centering
\small
\begin{tabular}{@{}p{0.94\columnwidth}@{}}
\toprule
\textbf{CSP} \\
\midrule
\ttfamily\footnotesize You are a person with \{culture\} cultural background. You will be performing a sentence completion task. Please directly complete the sentence. \\
\midrule
\textbf{CCT} \\
\midrule
\ttfamily\footnotesize You are a person with \{culture\} cultural background. You will be performing a sentence completion task. Please directly complete the sentence. Before you respond, think about how \{culture\} culture is different from Western cultures. \\
\bottomrule
\end{tabular}
\caption{Instruction prefixes for the two prompting baselines.
\texttt{\{culture\}} is filled with the cultural group of the setting. Both are prepended to the benchmark context, which is otherwise unchanged.}
\label{tab:prompt-templates}
\end{table}
 
Everything downstream is shared with the other methods. The prefix is prepended to each context string and evaluation then calls the same routine used for CoCoA and the baselines, with the same held-out folds, the same cached entity priors, the same up to $30 \times 30$ pairwise comparisons, and the same test-time PMI with $\alpha = 1.0$. The entity priors are context-free, so PMI scoring is unaffected by the prefix, and the only difference from the vanilla row is the instruction. Running the same script with an empty prefix reproduces the vanilla numbers, which we use as a check that the shared components are loaded identically.

\begin{table}[t]
\centering
\small
\begin{tabular}{llcc}
\toprule
Backbone & Method & $\mathrm{CBS}_{g}\downarrow$ & $\mathrm{CBS}_{n}\rightarrow 50$ \\
\midrule
\multirow{4}{*}{Llama-3.1-8B}
 & Vanilla & \score{36.4}{1.4} & \score{43.4}{1.3} \\
 & CSP     & \score{36.2}{1.5} & \score{40.5}{1.4} \\
 & CCT     & \score{36.9}{1.5} & \score{42.1}{1.6} \\
 & CoCoA   & \score{15.2}{1.3} & \score{50.3}{0.9} \\
\midrule
\multirow{4}{*}{Qwen3-8B}
 & Vanilla & \score{41.7}{1.2} & \score{45.8}{1.2} \\
 & CSP     & \score{41.1}{1.2} & \score{43.3}{1.1} \\
 & CCT     & \score{41.7}{1.3} & \score{43.4}{1.2} \\
 & CoCoA   & \score{24.9}{1.6} & \score{50.7}{0.8} \\
\bottomrule
\end{tabular}
\caption{Culture-aware prompting against CoCoA, macro-averaged over the ten
language settings under 5-fold cross-validation and reported as mean$\pm$SE
over the 50 setting--fold runs.}
\label{tab:prompting_base}
\end{table}

\subsection{Joint Multicultural Adapter}
\label{app:joint}
 
CoCoA trains one adapter per language setting, so a deployed system needs to know which setting an input belongs to. We therefore ask whether a single adapter can serve all ten. We pool the training data of the ten settings and train one adapter on it, keeping the objective, pairing, epochs, and default weights unchanged, and evaluate it on the same held-out folds.
 
\begin{table}[t]
\centering
\small
\begin{tabular}{lccc}
\toprule
Training scheme & $\mathrm{CBS}_{g}\downarrow$ & $\mathrm{CBS}_{n}\rightarrow 50$ \\
\midrule
Setting-specific & \score{15.2}{1.3} & \score{50.3}{0.9} \\
Single joint     & \score{17.2}{1.4} & \score{56.2}{1.7} \\
\midrule
Vanilla          & \score{36.4}{1.4} & \score{43.4}{1.3} \\
\bottomrule
\end{tabular}
\caption{One adapter trained on the pooled data of all ten language settings, against the setting-specific adapters used elsewhere in the paper. Llama-3.1-8B, macro-averaged over the ten settings under 5-fold cross-validation, reported as mean$\pm$SE over the 50 setting--fold runs.}
\label{tab:joint}
\end{table}
 
Table~\ref{tab:joint} reports the comparison. The grounded objective survives pooling. The joint adapter lowers $\mathrm{CBS}_{g}$ from 36.4 to 17.2, which recovers about 91\% of the reduction that setting-specific adapters achieve, and the remaining gap of 2.0 points is small next to the 21.2-point improvement itself. A single adapter is therefore a workable routing-free option when the grounded objective is what matters.
 
Neutral calibration is where pooling costs something. The joint adapter reaches $\mathrm{CBS}_{n}$ of 56.2 against 50.3 for setting-specific adapters, so it overshoots the balanced target by 6.2 points where the setting-specific adapters land within 0.3. The overshoot runs toward other-culture entities, while the vanilla model at 43.4 leans the other way by a comparable amount.
 
Part of this is weight calibration and not pooling. Appendix~\ref{app:weights} shows that per-setting adapters trained with the same default weights already reach 53.5, so a single global $w_d$ is mis-calibrated for several settings whether or not the data are pooled. The remainder, from 53.5 to 56.2, is not explained by the weights, since the joint run uses the same values, and is consistent with interference between settings that share one adapter. We do not separate the two effects further.

\begin{table}[!t]
\centering
\small
\setlength{\tabcolsep}{5pt}
\begin{tabular}{lcccc}
\toprule
\multirow{2}{*}[-0.7ex]{Method} & \textbf{CrowS-Pairs} & \multicolumn{3}{c}{\textbf{StereoSet}} \\
\cmidrule(lr){2-2} \cmidrule(lr){3-5}
 & Stereo. Pref.$\to50$ & SS$\to50$ & LMS$\uparrow$ & ICAT$\uparrow$ \\
\midrule
Vanilla & 65.8 & 66.40 & 92.53 & 62.19 \\
CoCoA   & 65.3 & 66.23 & 92.20 & 62.27 \\
\quad$\Delta$ 
        & \textcolor{gray}{\footnotesize$-$0.5} 
        & \textcolor{gray}{\footnotesize$-$0.17} 
        & \textcolor{gray}{\footnotesize$-$0.33} 
        & \textcolor{gray}{\footnotesize$+$0.08} \\
\bottomrule
\end{tabular}
\caption{Social bias evaluation on Llama-3.1-8B.
    CrowS-Pairs reports stereotypical preference rate, and StereoSet reports Stereotype Score (SS), Language Model Score (LMS), and ICAT. For Stereo. Pref. and SS, 50\% indicates no systematic stereotypical preference.}
\label{tab:social_bias}
\end{table}
\begin{table*}[t]
\centering
\small
\setlength{\tabcolsep}{5pt}
\renewcommand{\arraystretch}{1.08}
\begin{tabular}{@{}p{0.14\textwidth}p{0.10\textwidth}p{0.48\textwidth}r@{}}
\toprule
\textbf{Group} & \textbf{Symbol} & \textbf{Description} & \textbf{Value} \\
\midrule
\multirow{3}{*}{LoRA}
  & $r$ & Rank & $16$ \\
  & $\alpha_{\mathrm{LoRA}}$ & LoRA scaling factor & $32$ \\
  & $p_{\mathrm{drop}}$ & Dropout & $0.05$ \\
\midrule
\multirow{6}{*}{Optimization}
  & $\eta$ & Peak learning rate (AdamW) & $1\!\times\!10^{-5}$ \\
  & -- & Weight decay & $0.01$ \\
  & -- & Warmup ratio (cosine schedule) & $0.1$ \\
  & -- & Training epochs & $15$ \\
  & -- & Pairs per batch & $16$ \\
  & -- & Gradient clipping max norm & $1.0$ \\
\midrule
\multirow{2}{*}{PMI}
  & $\alpha_g,\,\alpha_n$ & Prior correction for training (grounded, neutral) & $1.0,\;0.3$ \\
  & $\alpha$ & Prior correction for evaluation & $1.0$ \\
\midrule
\multirow{4}{*}{Loss \& gradient}
  & $\tau$ & Contrastive temperature for $\mathcal{L}_g$ & $1.0$ \\
  & $\kappa,\,\lambda$ & Gap scaling for $\mathcal{L}_n$ and $\mathcal{L}_d$ & $10.0$ \\
  & $w_g,\,w_n,\,w_d$ & Objective weights & $1.0,\;2.0,\;1.0$ \\
\midrule
\multirow{1}{*}{Evaluation}
  & $K$ & Cross-validation folds & $5$ \\
\bottomrule
\end{tabular}
\caption{Default CoCoA hyperparameters (Llama-3.1-8B). LoRA targets attention projections ($W_Q, W_K, W_V, W_O$) across all layers. The vanilla model is kept as a fixed snapshot. Language-specific adjustments are in Table~\ref{tab:hparams}.}
\label{tab:cocoa-hparams}
\end{table*}
\subsection{Effect on Social Bias Metrics}
Recent work has shown that social bias mitigation can degrade cultural commonsense in LLMs~\cite{biasmitigationorculture}. 
This raises the reverse question of whether cultural alignment increases measured social bias. 
To examine this, we evaluate CoCoA-trained Llama-3.1-8B checkpoints on two widely used social bias benchmarks, CrowS-Pairs~\cite{crowspairs} and StereoSet~\cite{stereoset}. 
CrowS-Pairs reports the rate at which a model prefers stereotypical sentences over anti-stereotypical ones, where 50\% indicates no systematic preference. 
StereoSet reports Stereotype Score (SS), where 50\% is ideally debiased, along with Language Model Score (LMS) and ICAT, where higher values are better.

As shown in Table~\ref{tab:social_bias}, we observe no increase in measured social bias on either benchmark after CoCoA training. 
CrowS-Pairs decreases slightly from 65.8 to 65.3, and StereoSet SS changes from 66.40 to 66.23, with only marginal changes in LMS and ICAT. 
These results provide no evidence that context-conditional cultural alignment increases measured social bias on these benchmarks.

% ============================================
\section{Stability of Entity Priors}
\label{app:prior_drift}
\begin{table}[!t]
\centering
\small
\resizebox{\columnwidth}{!}{%
\begin{tabular}{llrrrrr}
\toprule
& & & \multicolumn{2}{c}{CBS$_g$ ($\downarrow$)} & \multicolumn{2}{c}{$|$CBS$_n - 50|$ ($\downarrow$)} \\
\cmidrule(lr){4-5} \cmidrule(lr){6-7}
Model & Language & Mean $|\Delta|$ & vanilla & CoCoA & vanilla & CoCoA \\
\midrule
\multirow{10}{*}{Llama-3.1-8B}
 & KO & 0.11 & 3.6  & 3.4  & 1.46  & 1.71  \\
 & JA & 0.51 & 6.3  & 5.0  & 4.74  & 4.40  \\
 & ZH & 0.99 & 14.1 & 14.4 & 2.67  & 4.87  \\
 & VI & 0.52 & 7.6  & 5.8  & 14.06 & 20.24 \\
 & UR & 0.24 & 11.6 & 11.2 & 2.81  & 3.38  \\
 & HI & 0.64 & 15.4 & 11.6 & 3.66  & 7.95  \\
 & ML & 0.14 & 28.6 & 27.8 & 5.73  & 3.93  \\
 & MR & 0.48 & 22.4 & 21.5 & 5.98  & 5.16  \\
 & GU & 0.55 & 30.3 & 28.7 & 3.57  & 3.75  \\
 & AR & 0.33 & 12.2 & 12.1 & 2.93  & 3.55  \\
\cmidrule{2-7}
 & \textbf{Avg} & \textbf{0.45} & \textbf{15.2} & \textbf{14.1} & \textbf{4.76} & \textbf{5.89} \\
\midrule
\multirow{10}{*}{Qwen3-8B}
 & KO & 0.30 & 6.7  & 6.2  & 4.90  & 4.88  \\
 & JA & 1.08 & 18.9 & 15.9 & 5.71  & 5.37  \\
 & ZH & 1.18 & 26.8 & 22.2 & 4.28  & 5.34  \\
 & VI & 1.94 & 17.9 & 11.7 & 6.44  & 16.28 \\
 & UR & 1.78 & 20.8 & 18.8 & 5.83  & 5.40  \\
 & HI & 1.97 & 30.5 & 27.0 & 6.00  & 3.84  \\
 & ML & 2.77 & 40.2 & 40.4 & 2.91  & 4.29  \\
 & MR & 2.04 & 36.0 & 35.5 & 4.58  & 5.69  \\
 & GU & 2.68 & 37.1 & 33.4 & 2.29  & 3.03  \\
 & AR & 1.89 & 14.6 & 13.6 & 2.83  & 1.75  \\
\cmidrule{2-7}
 & \textbf{Avg} & \textbf{1.76} & \textbf{24.9} & \textbf{22.5} & \textbf{4.58} & \textbf{5.59} \\
\bottomrule
\end{tabular}%
}
\caption{Prior drift and its impact on PMI-based CBS evaluation (5-fold average). Mean $|\Delta|$ measures the average absolute change in entity log-priors after LoRA fine-tuning. CBS columns compare evaluation with vanilla priors (used in all main experiments) against re-estimated CoCoA fine-tuned priors.}
\label{tab:prior_drift}
\end{table}

CoCoA uses PMI normalization (Eq.~\ref{eq:pmi}) with entity priors $\ell_{\mathrm{vanilla}}(e)$ estimated from the vanilla model before fine-tuning. A natural concern is whether LoRA adaptation shifts these priors enough to invalidate the PMI computation. If the fine-tuned model's marginal entity probabilities diverge substantially from the vanilla model's, the pre-computed priors would no longer reflect the correct frequency baseline.

\vspace{2mm}
\noindent\textbf{Methodology.}
For each CoCoA trained checkpoint, we re-estimate entity priors $\ell_{\mathrm{CoCoA}}(e)$ using the same BOS-only prompt protocol as the vanilla model. We then measure per-entity absolute drift $|\Delta_e| = |\ell_{\mathrm{CoCoA}}(e) - \ell_{\mathrm{vanilla}}(e)|$ and compare CBS computed with vanilla priors against CBS computed with fine-tuned priors. All numbers are averaged across the ten language settings and five folds.

\vspace{2mm}
\noindent\textbf{Results.}
Table~\ref{tab:prior_drift} reports the drift magnitude and its impact on CBS evaluation. For Llama-3.1-8B, the mean absolute drift is 0.45 log-probability units, while Qwen3-8B shows approximately 3.9$\times$ larger drift at 1.76 units. Despite this difference, the impact on CBS remains small for both models. Switching from vanilla to CoCoA fine-tuned priors changes the average CBS$_g$ by 1.1 percentage points for Llama (15.2 to 14.1) and 2.4 points for Qwen (24.9 to 22.5). The neutral-context deviation $|$CBS$_n - 50|$ shows similarly small changes, moving from 4.76 to 5.89 for Llama and from 4.58 to 5.59 for Qwen.

\vspace{2mm}
\noindent\textbf{Interpretation.}
CoCoA's LoRA adapters modify attention patterns to achieve context-conditional entity preference, but they do not substantially alter the model's unconditional entity priors. This is expected because LoRA operates on attention projection matrices, which govern how the model attends to contextual information, rather than MLP layers where entity-level knowledge is primarily stored~\cite{attentionmlp}. This stability justifies computing priors once from the vanilla model and reusing them throughout training and evaluation.

\section{Objective Design}
\label{app:objective}
 
\subsection{The Combined Neutral Target}
\label{app:derivation}
 
Section~\ref{sec:cocoa} states that $\mathcal{L}_n$ and $\mathcal{L}_d$ together define a single shrinkage target for the neutral score gap. We derive that target here.
 
Fix a neutral context $c_n$ and write $\Delta$ for the score gap $\Delta_\theta(c_n)$ and $\Delta_v$ for the vanilla gap $\Delta_{\mathrm{vanilla}}(c_n)$. The two neutral-side objectives are
\[
\mathcal{L}_n = \left(\frac{\Delta}{\kappa}\right)^{2},
\qquad
\mathcal{L}_d = \left(\frac{\Delta - \Delta_v}{\lambda}\right)^{2},
\]
and their weighted sum is
\[
J(\Delta)
= \frac{w_n}{\kappa^{2}}\,\Delta^{2}
+ \frac{w_d}{\lambda^{2}}\,(\Delta - \Delta_v)^{2}.
\]
For $w_n, w_d > 0$ this is a strictly convex quadratic in $\Delta$, since $J''(\Delta) = 2\,(w_n/\kappa^{2} + w_d/\lambda^{2}) > 0$, so it has a unique minimizer. Setting $J'(\Delta) = 0$ gives
\[
\Delta^{\star} = \rho\,\Delta_v,
\qquad
\rho = \frac{w_d/\lambda^{2}}{w_n/\kappa^{2} + w_d/\lambda^{2}}.
\]
 
Because $\rho$ lies strictly between zero and one for any positive weights, $\Delta^{\star}$ lies strictly between zero and the vanilla gap. The two terms therefore specify one shrinkage of the vanilla gap toward zero, and $\rho$ is the fraction of that gap which is retained. They enter only through the ratio $w_d\lambda^{-2}$ to $w_n\kappa^{-2}$, so $\kappa$ and $\lambda$ could be absorbed into $w_n$ and $w_d$ without changing the target. We keep them explicit to separate residual normalization from objective weighting.
 
The limiting cases recover the individual objectives. As $w_d \to 0$ we have $\rho \to 0$ and $\Delta^{\star} \to 0$, the balanced gap that $\mathcal{L}_n$ alone would enforce. As $w_n \to 0$ we have $\rho \to 1$ and $\Delta^{\star} \to \Delta_v$, the vanilla behavior that $\mathcal{L}_d$ alone would preserve.
 
This characterizes the target the two terms jointly specify for a fixed instance, and not the value that training reaches. The realized gap is a function of the adapter parameters rather than a free variable, $\mathcal{L}_g$ acts on those same parameters, and $\mathrm{CBS}_n$ counts pairwise comparisons rather than measuring the gap directly, so the correspondence to measured behavior is ordinal, not numerical. Table~\ref{tab:weight_sensitivity} shows the expected ordering. Removing $\mathcal{L}_d$ leaves only the anchor at zero and $\mathrm{CBS}_n$ rises to 58.8, removing $\mathcal{L}_n$ leaves only the anchor at the vanilla gap and $\mathrm{CBS}_n$ falls to 37.0, and the default configuration lies between the two at 53.3.

\subsection{Gradient Combination}
\label{app:gradient}
 
Section~\ref{sec:cocoa} states that the goal-aware projection step is a safeguard against interference and not the source of CoCoA's improvement. This section gives the comparison behind that statement.

We train CoCoA twice and vary only how the two gradient directions are combined. The first run uses the goal-aware projection of Eq.~\eqref{eq:pcgrad_beta} through Eq.~\eqref{eq:pcgrad_final}. The second replaces it with a plain weighted sum, $\mathbf{v}_f = \mathbf{v}_g + \mathbf{v}_n$, so no projection is ever applied. Losses, objective weights, data splits, pairing, epochs, and optimizer settings are identical in both runs.

\begin{table}[t]
\centering
\small
\begin{tabular}{lcc}
\toprule
Gradient combination & $\mathrm{CBS}_{g}\downarrow$ & $\mathrm{CBS}_{n}\rightarrow 50$ \\
\midrule
Goal-aware projection & \score{15.2}{1.3} & \score{50.3}{0.9} \\
Weighted sum          & \score{15.3}{1.3} & \score{49.9}{0.9} \\
\midrule
Vanilla               & \score{36.4}{1.4} & \score{43.4}{1.3} \\
\bottomrule
\end{tabular}
\caption{Gradient combination on Llama-3.1-8B, macro-averaged over the ten language settings under 5-fold cross-validation, reported as mean$\pm$SE over the 50 setting--fold runs. The two CoCoA rows use the same objective weights and differ only in whether the projection is applied.}
\label{tab:gradient}
\end{table}
 
Table~\ref{tab:gradient} reports both runs. They differ by 0.1 in $\mathrm{CBS}_{g}$ and 0.4 in $\mathrm{CBS}_{n}$, both far inside the reported standard errors, and neither is better.

The per-setting weights were selected in runs that used the projection, so this comparison holds them at values chosen under one of the two conditions. The projection nonetheless shows no advantage.

Both runs reduce $\mathrm{CBS}_{g}$ from 36.4 to roughly 15 and bring $\mathrm{CBS}_{n}$ from 43.4 to within half a point of 50, so the context-conditional behavior comes from the dual-context objective rather than from the gradient surgery. We retain the projection because it is inexpensive and guards against interference in configurations we have not swept, but the results we report do not depend on it.

\section{Behavioral Analysis}
\label{app:behavior}
 
\subsection{Score Decomposition}
\label{app:decomposition}
 
Section~\ref{sec:analysis} compares the local--Western score gap before and after training. A gap can widen because the local entity is scored higher, because the other-culture entity is scored lower, or both, and Figure~\ref{fig:per_pair} does not separate these. We therefore decompose the change into the two sides.
 
For each entity and context we take the difference between the prior-adjusted score under CoCoA and under the vanilla model. The same cached vanilla prior is subtracted on both sides of that difference, so it cancels, and the quantity reported below is the change in completion log-probability.
 
\begin{table}[t]
\centering
\small
\begin{tabular}{lccc}
\toprule
Context & $\Delta$ local & $\Delta$ other-culture & $\Delta$ gap \\
\midrule
Grounded & $+0.21$ & $-6.87$ & $+7.08$ \\
Neutral  & $+0.38$ & $+1.26$ & $-0.88$ \\
\bottomrule
\end{tabular}
\caption{Change in prior-adjusted score from the vanilla model to CoCoA, in
log-probability units, on Llama-3.1-8B over the ten language settings and five
folds. Positive values mean the entity is scored higher after training.}
\label{tab:decomposition}
\end{table}
 
Table~\ref{tab:decomposition} reports the decomposition. In grounded contexts the change is almost entirely on the other-culture side. The mean other-culture score falls by 6.87 while the local score rises by 0.21, so the widened gap reflects a Western entity being ruled out rather than a local entity being promoted. Local scores rise in 54.8\% of grounded observations, close to chance.
 
Three observations indicate that this is re-ranking conditioned on the cue and not a general aversion to Western entities. First, the decrease occurs only where the benchmark's explicit cue makes the other-culture entity inappropriate. In ``a representative \emph{Korean} dessert drink \_\_'', espresso is not an appropriate completion, and $\mathrm{CBS}_{g} \rightarrow 0$ requires that it not outrank sikhye there. Second, in neutral contexts the other-culture entity is not suppressed and in fact gains more than the local entity, $+1.26$ against $+0.38$. This is the score-level counterpart of $\mathrm{CBS}_{n}$ moving from 43.4 to 50.3, since the vanilla model under-ranks other-culture entities in cue-free contexts and CoCoA corrects in their direction. CBS counts pairwise orderings, so a gap change of under one log-probability unit moves it by several points when the neutral gaps are concentrated near zero. Third, the effect does not extend past the training objective, as general benchmark performance is essentially unchanged (Table~\ref{tab:general_perf}) and the social bias benchmarks do not move (Table~\ref{tab:social_bias}). A broad aversion to Western entities would be expected to appear in at least one of these.
 
The same decomposition bears on whether CoCoA amplifies local-culture associations. Local entities barely move in grounded contexts, so the model is not learning that Korean speakers always drink sikhye. It is learning that espresso is inappropriate when a Korean drink is explicitly requested.
 
\subsection{Entity-Level Context Sensitivity}
\label{app:consistency}
 
The results above are averages over entity pairs, which leaves open whether the context-conditional pattern holds for individual entities or is carried by a subset of pairings. Each local entity is paired with many Western counterparts under the $N \times N$ strategy, so we can check it entity by entity.
 
For a context type $t \in \{g, n\}$ we write
\[
\Delta_t = \Delta_{\theta}(c_t) - \Delta_{\mathrm{vanilla}}(c_t),
\]
where $\Delta_{\theta}$ is the score gap of Eq.~(\ref{eq:gap}) under CoCoA and $\Delta_{\mathrm{vanilla}}$ the same gap under the vanilla model. $\Delta_t$ is the change in the local--other gap under context type $t$, so $\Delta_g > \Delta_n$ says that CoCoA moved the pair further toward the local entity when a cue was present than when it was not.
 
\begin{table}[t]
\centering
\small
\begin{tabular}{@{}>{\raggedright\arraybackslash}p{0.62\columnwidth}r@{}}
\toprule
Held-out entity-level result & Value \\
\midrule
Pairs with $\Delta_g > 0$        & \score{93.8}{0.8}\% \\
Pairs with $\Delta_g > \Delta_n$ & \score{95.6}{0.6}\% \\
Entities with $\Delta_g > \Delta_n$ for $\geq$80\% of counterparts
                                 & 83.1\% \\
\bottomrule
\end{tabular}
\caption{Context sensitivity measured over held-out local entities and all their Western counterparts, on Llama-3.1-8B, with mean$\pm$SE over the 50 setting--fold runs.}
\label{tab:consistency}
\end{table}
 
Table~\ref{tab:consistency} reports the entity-level proportions. The pattern is not confined to a few pairings. Averaged over entities, $\Delta_g > \Delta_n$ holds for 88.5\% of a given entity's counterparts, and 83.1\% of entities satisfy it for at least four counterparts in five. The separation is also substantial. Over this narrower set of held-out entities and their counterparts, the mean gap shift is $+6.95$ when a cue is present and $-0.60$ when it is absent, and the small negative value in neutral contexts is the correction toward balance that accompanies $\mathrm{CBS}_{n}$ moving from 43.4 to 50.3.
 
These are proportions over held-out entities and their counterparts, and we do not claim a formal guarantee that the ordering holds for every pair.

% Appendix: Remaining 5 cultures — same format as main table

\begin{table*}[p]
\centering
\setlength{\tabcolsep}{2.2pt}
\renewcommand{\arraystretch}{1.05}
\resizebox{\textwidth}{!}{%
\begin{tabular}{l | ccc c | ccc c | ccc c | ccc c | ccc c | ccc c}
\toprule
\multicolumn{25}{c}{\textbf{Llama-3.1-8B}} \\
\midrule
& \multicolumn{4}{c|}{\textsc{Chinese}}
& \multicolumn{4}{c|}{\textsc{Urdu}}
& \multicolumn{4}{c|}{\textsc{Malayalam}}
& \multicolumn{4}{c|}{\textsc{Marathi}}
& \multicolumn{4}{c|}{\textsc{Gujarati}}
& \multicolumn{4}{c}{\textcolor{red}{\textsc{Avg}}} \\
\cmidrule(lr){2-5}\cmidrule(lr){6-9}\cmidrule(lr){10-13}\cmidrule(lr){14-17}\cmidrule(lr){18-21}\cmidrule(lr){22-25}
Method
& \CBSg$\downarrow$ & {\scriptsize$\Delta g$} & \CBSn$\!\to\!50$ & CS$\uparrow$
& \CBSg$\downarrow$ & {\scriptsize$\Delta g$} & \CBSn$\!\to\!50$ & CS$\uparrow$
& \CBSg$\downarrow$ & {\scriptsize$\Delta g$} & \CBSn$\!\to\!50$ & CS$\uparrow$
& \CBSg$\downarrow$ & {\scriptsize$\Delta g$} & \CBSn$\!\to\!50$ & CS$\uparrow$
& \CBSg$\downarrow$ & {\scriptsize$\Delta g$} & \CBSn$\!\to\!50$ & CS$\uparrow$
& \CBSg$\downarrow$ & {\scriptsize$\Delta g$} & \CBSn$\!\to\!50$ & CS$\uparrow$ \\
\midrule
Vanilla
  & \score{36.2}{3.7} & -- & \score{43.1}{1.8} & 6.9
  & \underline{\score{29.5}{4.2}} & -- & \score{44.6}{3.1} & 15.1
  & \score{44.5}{2.8} & -- & \underline{\score{51.5}{5.5}} & \underline{7.0}
  & \underline{\score{43.6}{1.1}} & -- & \underline{\score{51.2}{0.6}} & \underline{7.6}
  & \underline{\score{44.1}{1.9}} & -- & \underline{\score{49.4}{2.6}} & \underline{5.3}
  & \underline{\score{39.6}{2.7}} & -- & \underline{\score{48.0}{1.7}} & 8.4 \\
BiasUnlearn
  & \score{39.5}{3.1} & {\scriptsize+3.3} & \underline{\score{47.6}{3.0}} & \underline{8.1}
  & \score{31.4}{3.6} & {\scriptsize+1.9} & \textbf{\score{49.0}{3.6}} & \underline{17.6}
  & \underline{\score{44.4}{3.3}} & {\scriptsize$-$0.1} & \textbf{\score{51.2}{7.4}} & 6.8
  & \underline{\score{43.6}{1.1}} & {\scriptsize+0.0} & \textbf{\score{51.1}{0.5}} & 7.5
  & \score{44.3}{1.8} & {\scriptsize+0.2} & \textbf{\score{49.5}{2.9}} & 5.2
  & \score{40.6}{2.3} & {\scriptsize+1.0} & \textbf{\score{49.7}{0.6}} & \underline{9.0} \\
BiasEdit
  & \underline{\score{26.2}{2.0}} & {\scriptsize$-$10.0} & \score{26.7}{1.6} & 0.5
  & \score{41.5}{5.3} & {\scriptsize+12.0} & \score{40.8}{3.6} & 0.7
  & \score{46.3}{2.3} & {\scriptsize+1.8} & \score{42.8}{4.6} & 3.5
  & \score{44.1}{1.7} & {\scriptsize+0.5} & \score{43.5}{1.9} & 0.6
  & \score{44.5}{1.4} & {\scriptsize+0.4} & \score{41.6}{2.6} & 2.9
  & \score{40.5}{3.6} & {\scriptsize+0.9} & \score{39.1}{2.9} & 1.6 \\
\textbf{CoCoA}
  & \textbf{\score{14.1}{3.9}} & {\scriptsize$-$22.1} & \textbf{\score{49.3}{3.2}} & \textbf{35.2}
  & \textbf{\score{11.6}{4.0}} & {\scriptsize$-$17.9} & \underline{\score{52.4}{4.1}} & \textbf{40.8}
  & \textbf{\score{28.6}{3.0}} & {\scriptsize$-$15.9} & \score{52.0}{7.1} & \textbf{23.4}
  & \textbf{\score{22.4}{2.9}} & {\scriptsize$-$21.2} & \score{54.7}{4.5} & \textbf{32.3}
  & \textbf{\score{30.3}{3.6}} & {\scriptsize$-$13.8} & \score{53.6}{3.3} & \textbf{23.3}
  & \textbf{\score{21.4}{3.8}} & {\scriptsize$-$18.2} & \score{52.4}{0.9} & \textbf{31.0} \\
  
\midrule
\multicolumn{25}{c}{\textbf{Qwen3-8B}} \\
\midrule
& \multicolumn{4}{c|}{\textsc{Chinese}}
& \multicolumn{4}{c|}{\textsc{Urdu}}
& \multicolumn{4}{c|}{\textsc{Malayalam}}
& \multicolumn{4}{c|}{\textsc{Marathi}}
& \multicolumn{4}{c|}{\textsc{Gujarati}}
& \multicolumn{4}{c}{\textcolor{red}{\textsc{Avg}}} \\
\cmidrule(lr){2-5}\cmidrule(lr){6-9}\cmidrule(lr){10-13}\cmidrule(lr){14-17}\cmidrule(lr){18-21}\cmidrule(lr){22-25}
Method
& \CBSg$\downarrow$ & {\scriptsize$\Delta g$} & \CBSn$\!\to\!50$ & CS$\uparrow$
& \CBSg$\downarrow$ & {\scriptsize$\Delta g$} & \CBSn$\!\to\!50$ & CS$\uparrow$
& \CBSg$\downarrow$ & {\scriptsize$\Delta g$} & \CBSn$\!\to\!50$ & CS$\uparrow$
& \CBSg$\downarrow$ & {\scriptsize$\Delta g$} & \CBSn$\!\to\!50$ & CS$\uparrow$
& \CBSg$\downarrow$ & {\scriptsize$\Delta g$} & \CBSn$\!\to\!50$ & CS$\uparrow$
& \CBSg$\downarrow$ & {\scriptsize$\Delta g$} & \CBSn$\!\to\!50$ & CS$\uparrow$ \\
\midrule
Vanilla
  & \score{39.2}{2.1} & -- & \score{42.7}{3.6} & 3.5
  & \score{37.3}{3.4} & -- & \score{43.0}{3.8} & 5.7
  & \score{51.8}{3.0} & -- & \score{56.0}{2.1} & \underline{4.2}
  & \score{49.1}{2.6} & -- & \score{54.9}{1.9} & \underline{5.8}
  & \score{43.7}{2.9} & -- & \textbf{\score{49.5}{2.9}} & \underline{5.8}
  & \score{44.2}{2.7} & -- & \score{49.2}{2.8} & \underline{5.0} \\
BiasUnlearn
  & \score{40.3}{2.0} & {\scriptsize+1.1} & \underline{\score{44.2}{3.9}} & \underline{3.9}
  & \score{40.9}{6.1} & {\scriptsize+3.6} & \underline{\score{47.2}{6.1}} & \underline{6.3}
  & \score{48.9}{4.0} & {\scriptsize$-$2.9} & \underline{\score{52.4}{3.1}} & 3.5
  & \score{47.1}{3.8} & {\scriptsize$-$2.0} & \textbf{\score{52.6}{1.8}} & 5.5
  & \score{43.8}{3.3} & {\scriptsize+0.1} & \underline{\score{49.5}{3.4}} & 5.7
  & \score{44.2}{1.7} & {\scriptsize+0.0} & \underline{\score{49.2}{1.6}} & \underline{5.0} \\
BiasEdit
  & \textbf{\score{25.2}{1.3}} & {\scriptsize$-$14.0} & \score{25.9}{2.2} & 0.7
  & \underline{\score{35.9}{3.0}} & {\scriptsize$-$1.4} & \score{35.4}{2.6} & 0.5
  & \underline{\score{43.4}{2.3}} & {\scriptsize$-$8.4} & \score{41.4}{3.0} & 2.0
  & \underline{\score{44.3}{4.2}} & {\scriptsize$-$4.8} & \score{44.0}{4.0} & 0.3
  & \underline{\score{42.1}{1.1}} & {\scriptsize$-$1.6} & \score{40.4}{1.6} & 1.7
  & \underline{\score{38.2}{3.4}} & {\scriptsize$-$6.0} & \score{37.4}{3.2} & 0.8 \\
\textbf{CoCoA}
  & \underline{\score{26.8}{6.1}} & {\scriptsize$-$12.4} & \textbf{\score{47.0}{5.1}} & \textbf{20.2}
  & \textbf{\score{20.8}{5.4}} & {\scriptsize$-$16.5} & \textbf{\score{51.9}{7.6}} & \textbf{31.1}
  & \textbf{\score{40.2}{3.9}} & {\scriptsize$-$11.6} & \textbf{\score{52.3}{3.3}} & \textbf{12.1}
  & \textbf{\score{36.0}{3.3}} & {\scriptsize$-$13.1} & \underline{\score{53.4}{4.0}} & \textbf{17.4}
  & \textbf{\score{37.1}{2.1}} & {\scriptsize$-$6.6} & \score{49.2}{2.4} & \textbf{12.1}
  & \textbf{\score{32.2}{3.5}} & {\scriptsize$-$12.0} & \textbf{\score{50.8}{1.2}} & \textbf{18.6} \\

\midrule
\multicolumn{25}{c}{\textbf{Gemma-3-12B-pt}} \\
\midrule
& \multicolumn{4}{c|}{\textsc{Chinese}}
& \multicolumn{4}{c|}{\textsc{Urdu}}
& \multicolumn{4}{c|}{\textsc{Malayalam}}
& \multicolumn{4}{c|}{\textsc{Marathi}}
& \multicolumn{4}{c|}{\textsc{Gujarati}}
& \multicolumn{4}{c}{\textcolor{red}{\textsc{Avg}}} \\
\cmidrule(lr){2-5}\cmidrule(lr){6-9}\cmidrule(lr){10-13}\cmidrule(lr){14-17}\cmidrule(lr){18-21}\cmidrule(lr){22-25}
Method
& \CBSg$\downarrow$ & {\scriptsize$\Delta g$} & \CBSn$\!\to\!50$ & CS$\uparrow$
& \CBSg$\downarrow$ & {\scriptsize$\Delta g$} & \CBSn$\!\to\!50$ & CS$\uparrow$
& \CBSg$\downarrow$ & {\scriptsize$\Delta g$} & \CBSn$\!\to\!50$ & CS$\uparrow$
& \CBSg$\downarrow$ & {\scriptsize$\Delta g$} & \CBSn$\!\to\!50$ & CS$\uparrow$
& \CBSg$\downarrow$ & {\scriptsize$\Delta g$} & \CBSn$\!\to\!50$ & CS$\uparrow$
& \CBSg$\downarrow$ & {\scriptsize$\Delta g$} & \CBSn$\!\to\!50$ & CS$\uparrow$ \\
\midrule
Vanilla
  & \score{52.8}{2.0} & -- & \score{52.4}{2.0} & 0.4
  & \score{48.9}{3.1} & -- & \score{51.3}{3.7} & \underline{2.4}
  & \score{48.0}{1.9} & -- & \underline{\score{48.5}{1.7}} & 0.5
  & \score{52.3}{3.2} & -- & \score{53.2}{2.8} & 0.9
  & \score{45.5}{2.0} & -- & \score{45.6}{3.4} & 0.1
  & \score{49.5}{1.3} & -- & \score{50.2}{1.4} & 0.9 \\
BiasUnlearn
  & \score{52.3}{2.6} & {\scriptsize$-$0.5} & \score{51.8}{2.3} & \underline{0.5}
  & \score{48.6}{3.4} & {\scriptsize$-$0.3} & \underline{\score{50.4}{4.8}} & 1.8
  & \score{48.0}{1.9} & {\scriptsize+0.0} & \textbf{\score{48.6}{1.6}} & 0.6
  & \score{51.9}{3.5} & {\scriptsize$-$0.4} & \score{52.7}{2.9} & 0.8
  & \score{45.7}{2.1} & {\scriptsize+0.2} & \underline{\score{45.9}{3.4}} & 0.2
  & \score{49.3}{1.2} & {\scriptsize$-$0.2} & \underline{\score{49.9}{1.2}} & 0.8 \\
BiasEdit
  & \underline{\score{51.7}{2.3}} & {\scriptsize$-$1.1} & \underline{\score{51.3}{2.3}} & 0.4
  & \underline{\score{47.9}{3.2}} & {\scriptsize$-$1.0} & \textbf{\score{50.1}{3.1}} & 2.2
  & \underline{\score{46.7}{2.3}} & {\scriptsize$-$1.3} & \score{47.5}{2.0} & \underline{0.8}
  & \underline{\score{51.3}{3.5}} & {\scriptsize$-$1.0} & \underline{\score{52.6}{2.9}} & \underline{1.3}
  & \underline{\score{44.5}{2.4}} & {\scriptsize$-$1.0} & \score{45.5}{3.3} & \underline{1.0}
  & \underline{\score{48.4}{1.4}} & {\scriptsize$-$1.1} & \score{49.4}{1.3} & \underline{1.1} \\
\textbf{CoCoA}
  & \textbf{\score{39.4}{2.0}} & {\scriptsize$-$13.4} & \textbf{\score{50.9}{1.8}} & \textbf{11.5}
  & \textbf{\score{34.1}{1.9}} & {\scriptsize$-$14.8} & \score{52.1}{3.7} & \textbf{18.0}
  & \textbf{\score{41.2}{2.5}} & {\scriptsize$-$6.8} & \score{48.1}{2.0} & \textbf{6.9}
  & \textbf{\score{41.0}{4.7}} & {\scriptsize$-$11.3} & \textbf{\score{52.2}{3.0}} & \textbf{11.2}
  & \textbf{\score{40.3}{3.4}} & {\scriptsize$-$5.2} & \textbf{\score{46.0}{3.1}} & \textbf{5.7}
  & \textbf{\score{39.2}{1.2}} & {\scriptsize$-$10.3} & \textbf{\score{49.9}{1.2}} & \textbf{10.7} \\

\midrule
\multicolumn{25}{c}{\textbf{Mistral-7B-v0.3}} \\
\midrule
& \multicolumn{4}{c|}{\textsc{Chinese}}
& \multicolumn{4}{c|}{\textsc{Urdu}}
& \multicolumn{4}{c|}{\textsc{Malayalam}}
& \multicolumn{4}{c|}{\textsc{Marathi}}
& \multicolumn{4}{c|}{\textsc{Gujarati}}
& \multicolumn{4}{c}{\textcolor{red}{\textsc{Avg}}} \\
\cmidrule(lr){2-5}\cmidrule(lr){6-9}\cmidrule(lr){10-13}\cmidrule(lr){14-17}\cmidrule(lr){18-21}\cmidrule(lr){22-25}
Method
& \CBSg$\downarrow$ & {\scriptsize$\Delta g$} & \CBSn$\!\to\!50$ & CS$\uparrow$
& \CBSg$\downarrow$ & {\scriptsize$\Delta g$} & \CBSn$\!\to\!50$ & CS$\uparrow$
& \CBSg$\downarrow$ & {\scriptsize$\Delta g$} & \CBSn$\!\to\!50$ & CS$\uparrow$
& \CBSg$\downarrow$ & {\scriptsize$\Delta g$} & \CBSn$\!\to\!50$ & CS$\uparrow$
& \CBSg$\downarrow$ & {\scriptsize$\Delta g$} & \CBSn$\!\to\!50$ & CS$\uparrow$
& \CBSg$\downarrow$ & {\scriptsize$\Delta g$} & \CBSn$\!\to\!50$ & CS$\uparrow$ \\
\midrule
Vanilla
  & \score{40.9}{2.5} & -- & \underline{\score{44.7}{3.3}} & 3.8
  & \underline{\score{40.9}{2.2}} & -- & \underline{\score{46.1}{3.6}} & 5.2
  & \score{56.1}{1.0} & -- & \underline{\score{52.7}{2.8}} & 3.4
  & \score{54.6}{3.1} & -- & \score{54.9}{2.9} & 0.3
  & \score{60.1}{2.0} & -- & \score{57.9}{3.7} & 2.2
  & \score{50.5}{4.0} & -- & \score{51.3}{2.5} & 3.0 \\
BiasUnlearn
  & \score{42.7}{4.0} & {\scriptsize+1.8} & \textbf{\score{46.7}{5.6}} & \underline{4.0}
  & \score{43.0}{3.8} & {\scriptsize+2.1} & \textbf{\score{48.4}{5.0}} & \underline{5.4}
  & \score{53.9}{2.5} & {\scriptsize$-$2.2} & \textbf{\score{49.9}{4.2}} & \underline{4.0}
  & \score{50.9}{3.5} & {\scriptsize$-$3.7} & \textbf{\score{51.1}{3.4}} & 0.2
  & \score{53.5}{8.0} & {\scriptsize$-$6.6} & \underline{\score{49.8}{8.7}} & \textbf{3.7}
  & \score{48.8}{2.4} & {\scriptsize$-$1.7} & \textbf{\score{49.2}{0.7}} & \underline{3.5} \\
BiasEdit
  & \underline{\score{29.3}{2.8}} & {\scriptsize$-$11.6} & \score{29.0}{2.5} & 0.3
  & \score{42.7}{3.7} & {\scriptsize+1.8} & \score{43.1}{3.0} & 0.4
  & \underline{\score{47.1}{2.9}} & {\scriptsize$-$9.0} & \score{42.9}{4.6} & \textbf{4.2}
  & \underline{\score{42.6}{2.5}} & {\scriptsize$-$12.0} & \score{40.5}{2.3} & \underline{2.1}
  & \underline{\score{50.6}{2.8}} & {\scriptsize$-$9.5} & \score{47.9}{4.3} & \underline{2.7}
  & \underline{\score{42.5}{3.6}} & {\scriptsize$-$8.0} & \score{40.7}{3.1} & 1.9 \\
\textbf{CoCoA}
  & \textbf{\score{18.2}{4.8}} & {\scriptsize$-$22.7} & \score{55.4}{3.5} & \textbf{37.2}
  & \textbf{\score{14.0}{3.9}} & {\scriptsize$-$26.9} & \score{54.0}{6.7} & \textbf{40.0}
  & \textbf{\score{44.6}{3.6}} & {\scriptsize$-$11.5} & \score{44.4}{2.1} & 0.2
  & \textbf{\score{30.7}{2.3}} & {\scriptsize$-$23.9} & \underline{\score{52.0}{2.5}} & \textbf{21.3}
  & \textbf{\score{47.9}{2.9}} & {\scriptsize$-$12.2} & \textbf{\score{49.9}{3.1}} & 2.0
  & \textbf{\score{31.1}{6.6}} & {\scriptsize$-$19.4} & \underline{\score{51.1}{1.9}} & \textbf{20.1} \\
\bottomrule
\end{tabular}}
\caption{Results for the remaining five language settings (ZH, UR, ML, MR, GU). Same evaluation protocol and notation as Table~\ref{tab:main}. Avg is computed over these five settings only.}
\label{tab:main_rest}
\end{table*}
\clearpage
% Data Statistics (B6)

\begin{table*}[p]
\centering
\small
\setlength{\tabcolsep}{2.5pt}
\resizebox{\textwidth}{!}{%
\begin{tabular}{ll|cc|cc|cc|cc|cc|cc|cc|rr}
\toprule
& & \multicolumn{2}{c|}{\textbf{Auth}} & \multicolumn{2}{c|}{\textbf{Bev}} & \multicolumn{2}{c|}{\textbf{Food}} & \multicolumn{2}{c|}{\textbf{Loc}} & \multicolumn{2}{c|}{\textbf{Na(M)}} & \multicolumn{2}{c|}{\textbf{Na(F)}} & \multicolumn{2}{c|}{\textbf{Sport}} & \multicolumn{2}{c}{\textbf{Total}} \\
\cmidrule(lr){3-4}\cmidrule(lr){5-6}\cmidrule(lr){7-8}\cmidrule(lr){9-10}\cmidrule(lr){11-12}\cmidrule(lr){13-14}\cmidrule(lr){15-16}\cmidrule(lr){17-18}
Dataset & Language & G & N & G & N & G & N & G & N & G & N & G & N & G & N & G & N \\
\midrule
\multirow{9}{*}{Camellia}
 & Korean (KO)     & 31 & 42 & 30 & 52 & 34 & 64 & 26 & 24 & 29 & 37 & 29 & 37 & 29 & 26 & 179 & 245 \\
 & Japanese (JA)   & 32 & 33 & 26 & 27 & 29 & 29 & 27 & 27 & 19 & 19 & 19 & 19 & 23 & 24 & 156 & 159 \\
 & Chinese (ZH)    & 25 & 21 & 27 & 27 & 31 & 31 & 27 & 27 & 27 & 24 & 27 & 24 & 21 & 20 & 158 & 150 \\
 & Vietnamese (VI) & 18 & 32 & 14 & 39 & 67 & 46 & 33 & 44 & 16 & 14 & 21 &  7 & 10 & 16 & 179 & 198 \\
 & Urdu (UR)       & 10 & 10 & 10 & 10 & 10 & 10 & 10 & 10 & 10 & 10 & 10 & 10 & 10 & 10 &  70 &  70 \\
 & Hindi (HI)      & 24 & 23 & 23 & 22 & 39 & 35 & 42 & 36 & 28 & 26 & 26 & 27 & 33 & 23 & 215 & 192 \\
 & Malayalam (ML)  & 24 & 23 & 23 & 22 & 39 & 35 & 42 & 36 & 28 & 26 & 26 & 27 & 33 & 23 & 215 & 192 \\
 & Marathi (MR)    & 24 & 23 & 23 & 22 & 37 & 35 & 42 & 36 & 28 & 24 & 26 & 22 & 33 & 23 & 213 & 185 \\
 & Gujarati (GU)   & 24 & 23 & 23 & 22 & 39 & 35 & 42 & 36 & 28 & 26 & 26 & 27 & 33 & 23 & 215 & 192 \\
\midrule
CAMeL & Arabic (AR) & 22 & 42 & 22 & 52 & 23 & 65 & 37 & 25 & 37 & 49 & 40 & 46 & 28 & 39 & 209 & 318 \\
\bottomrule
\end{tabular}%
}
\caption{Number of grounded (G) and neutral (N) contexts per language and entity type. For KO, JA, and ZH, Names contexts are shared across male and female entity types. Indian cultures (HI, ML, MR, GU) share context sets with minor variations.}
\label{tab:ctx_stats}
\end{table*}

\begin{table*}[t]
\centering
\small
\setlength{\tabcolsep}{3.5pt}
\resizebox{\textwidth}{!}{%
\begin{tabular}{ll|rrrrrrr|rr}
\toprule
& & \multicolumn{7}{c|}{\textbf{Local Entities per Type}} & \multicolumn{2}{c}{\textbf{Total}} \\
\cmidrule(lr){3-9}\cmidrule(lr){10-11}
Dataset & Language & Auth & Bev & Food & Loc & Na(M) & Na(F) & Sport & Local & Other \\
\midrule
\multirow{9}{*}{Camellia}
 & Korean (KO)     & 602 & 107 & 416 & 1,260 & 899 & 886 & 266 & 4,436 & 3,709 \\
 & Japanese (JA)   & 260 & 115 & 635 & 817   & 503 & 523 & 354 & 3,207 & 3,699 \\
 & Chinese (ZH)    & 165 & 189 & 415 & 1,000 & 906 & 1,123 & 1,578 & 5,376 & 3,709 \\
 & Vietnamese (VI) & 24  & 89  & 376 & 92    & 251 & 151 & 51  & 1,034 & 3,709 \\
 & Urdu (UR)       & 43  & 10  & 74  & 196   & 334 & 163 & 17  & 837   & 2,939 \\
 & Hindi (HI)      & 207 & 34  & 605 & 181   & 651 & 563 & 165 & 2,406 & 2,939 \\
 & Malayalam (ML)  & 207 & 34  & 605 & 181   & 651 & 563 & 165 & 2,406 & 2,939 \\
 & Marathi (MR)    & 207 & 34  & 605 & 181   & 651 & 563 & 165 & 2,406 & 2,939 \\
 & Gujarati (GU)   & 207 & 34  & 603 & 181   & 648 & 561 & 165 & 2,399 & 2,939 \\
\midrule
CAMeL & Arabic (AR) & 207 & 54 & 326 & 1,061 & 340 & 537 & 1,270 & 3,795 & 13,999 \\
\bottomrule
\end{tabular}%
}
\caption{Number of local entities per language and entity type. Other-culture (Western) entities are shared within each benchmark. Indian cultures (HI, ML, MR, GU) share entity sets with minor variations.}
\label{tab:ent_stats}
\end{table*}
\clearpage
\begin{table*}[!ht]
\centering
\small
\setlength{\tabcolsep}{12pt}
\begin{tabular}{llll}
\toprule
\textbf{Artifact} & \textbf{Type} & \textbf{License} & \textbf{Source} \\
\midrule
Llama-3.1-8B     & Model   & Llama 3.1 Community & \href{https://huggingface.co/meta-llama/Llama-3.1-8B}{HuggingFace} \\
Qwen3-8B        & Model   & Apache 2.0          & \href{https://huggingface.co/Qwen/Qwen3-8B}{HuggingFace} \\
Mistral-7B-v0.3  & Model   & Apache 2.0          & \href{https://huggingface.co/mistralai/Mistral-7B-v0.3}{HuggingFace} \\
Gemma-3-12B-pt   & Model   & Gemma Terms of Use  & \href{https://huggingface.co/google/gemma-3-12b-pt}{HuggingFace} \\
\midrule
Camellia          & Dataset & MIT           & \href{https://github.com/tareknaous/camellia}{GitHub} \\
CAMeL             & Dataset & MIT           & \href{https://github.com/tareknaous/camel}{GitHub} \\
\midrule 
MMLU & \multirow{5}{*}{Library} & \multirow{5}{*}{MIT} & \multirow{5}{*}{\href{https://github.com/EleutherAI/lm-evaluation-harness}{lm-eval-harness}} \\ HellaSwag & & & \\ ARC-Challenge & & & \\ WinoGrande & & & \\ TruthfulQA & & & \\
\midrule 
CrowS-Pairs & Benchmark & CC-BY-SA 4.0 & \href{https://github.com/nyu-mll/crows-pairs}{GitHub} \\ 
StereoSet & Benchmark & CC-BY-SA 4.0 & \href{https://github.com/moinnadeem/StereoSet}{GitHub} \\
\bottomrule
\end{tabular}
\caption{Licenses and sources for all artifacts used in this work. All artifacts are publicly available and used consistently with their intended research purposes.}
\label{tab:licenses}
\end{table*}
\clearpage
\begin{table*}[!ht]
\centering
\small
\setlength{\tabcolsep}{5pt}
\renewcommand{\arraystretch}{1.08}
\label{tab:notation}
\begin{tabular}{@{}p{0.22\textwidth}p{0.73\textwidth}@{}}
\toprule
\textbf{Symbol} & \textbf{Description} \\
\midrule
\multicolumn{2}{@{}l}{\textit{Contexts and entities}} \\
$c_g,\ c_n$ & Grounded and neutral context strings \\
$\mathcal{C}_g,\ \mathcal{C}_n$ & Sets of grounded and neutral contexts \\
$e_l,\ e_o$ & Local entity and other-culture entity \\
$(e_l, e_o)$ & Culturally contrasting entity pair \\
\midrule
\multicolumn{2}{@{}l}{\textit{Scoring}} \\
$\theta = \{W, \phi\}$ & Full model parameters (frozen base $W$ + trainable LoRA adapters $\phi$) \\
$\ell_\theta(e \mid c)$ & Completion log-probability of entity $e$ given context $c$ \\
$\ell_{\mathrm{vanilla}}(e)$ & Entity prior from the vanilla model (BOS-only context) \\
$\alpha$ & PMI prior correction strength \\
$\alpha_g,\ \alpha_n$ & Context-specific prior correction for training (grounded, neutral) \\
$s_\theta(e \mid c)$ & Prior-adjusted score: $\ell_\theta(e \mid c) - \alpha \cdot \ell_{\mathrm{vanilla}}(e)$ \\
$\Delta_\theta(c)$ & Score gap: $s_\theta(e_l \mid c) - s_\theta(e_o \mid c)$ \\
$\Delta_{\mathrm{vanilla}}(c)$ & Score gap under the vanilla model \\
\midrule
\multicolumn{2}{@{}l}{\textit{Training objectives}} \\
$\mathcal{L}_g$ & Grounded alignment loss (contrastive) \\
$\mathcal{L}_n$ & Neutral calibration loss (squared gap) \\
$\mathcal{L}_d$ & Drift regularization (penalizes deviation from vanilla neutral gap) \\
$w_g,\ w_n,\ w_d$ & Objective weights \\
$\tau$ & Contrastive temperature for $\mathcal{L}_g$ \\
$\kappa,\ \lambda$ & Scaling constants for $\mathcal{L}_n$ and $\mathcal{L}_d$ \\
\midrule
\multicolumn{2}{@{}l}{\textit{Goal-aware gradient optimization}} \\
$\mathbf{v}_g,\ \mathbf{v}_n$ & Gradient directions for grounded and neutral objectives \\
$\tilde{\mathbf{v}}_g,\ \tilde{\mathbf{v}}_n$ & Projected gradient directions after conflict resolution \\
$d_g,\ d_n$ & Normalized distances from CBS targets \\
$\beta_{i \leftarrow j}$ & Projection strength for direction $i$ under conflict with $j$ \\
$\mathbf{v}_f$ & Reconciled gradient direction \\
\midrule
\multicolumn{2}{@{}l}{\textit{Evaluation}} \\
$\mathrm{CBS}_g,\ \mathrm{CBS}_n$ & Cultural Bias Score on grounded and neutral contexts \\
$\Delta\mathrm{CBS}_g$ & CBS$_g$ change after training: $\mathrm{CBS}_g^{\text{CoCoA}} - \mathrm{CBS}_g^{\text{vanilla}}$ \\
$\mathrm{CS}$ & Context sensitivity: $|\mathrm{CBS}_n - \mathrm{CBS}_g|$ \\
\bottomrule
\end{tabular}
\caption{Mathematical notation used throughout the paper.}
\end{table*}

\end{document}